%% file: main.tex
\documentclass[10pt,twocolumn,letterpaper]{article}

\usepackage{wacv}              % To produce the CAMERA-READY version
\input{preamble}

\definecolor{wacvblue}{rgb}{0.21,0.49,0.74}
\usepackage[pagebackref,breaklinks,colorlinks,allcolors=wacvblue]{hyperref}

\usepackage{url}

\usepackage{algorithm}
\usepackage{algorithmic}
\usepackage{booktabs}
\usepackage{times}
\usepackage{epsfig}
\usepackage{amsmath}
\usepackage{amssymb}
\usepackage{caption}
\usepackage{multirow}
\usepackage[flushleft]{threeparttable}
\usepackage{pifont}% http://ctan.org/pkg/pifont

\usepackage{color}
\usepackage{array}
\usepackage{makecell}

\usepackage{adjustbox} % in preamble

\newlength\savewidth
\usepackage[utf8]{inputenc} % allow utf-8 input
\usepackage[T1]{fontenc}    % use 8-bit T1 fonts
\usepackage{hyperref}       % hyperlinks
\usepackage{amsfonts}       % blackboard math symbols
\usepackage{nicefrac}       % compact symbols for 1/2, etc.
\usepackage{microtype}      % microtypography
\usepackage{xcolor}         % colors

\usepackage{tabularray}
\usepackage{colortbl}
\usepackage{arydshln}

\definecolor{mygray}{gray}{.9}
\definecolor{mygreen}{rgb}{0, 0.6, 0}
\definecolor{myteal}{rgb}{0.16, 0.47, 0.56}
\definecolor{mypink}{rgb}{0.81, 0.25, 0.44}

\usepackage{xr}

\def\wacvPaperID{2179} % *** Enter the WACV Paper ID here
\def\confName{WACV}
\def\confYear{2026}

\title{RobustGait: Robustness Analysis for Appearance Based Gait Recognition}

\author{
Reeshoon Sayera \quad Akash Kumar \quad Sirshapan Mitra \quad Prudvi Kamtam \quad Yogesh S Rawat \\
University of Central Florida\\
{\tt\small \{reeshon.sayera, akash.kumar, sirshapan.mitra, prudvi.kamtam, yogesh\}@ucf.edu} \\
\small
{\url{https://reeshoon.github.io/robustgaitbenchmark}}
}

\begin{document}
\maketitle
\input{sec/0_abstract}    
\input{sec/1_intro}

\input{sec/2_related_work}
\input{sec/3_benchmark_setup}

\input{sec/4_benchmarking_analysis}

\input{sec/5_improving_robustness}

\input{sec/6_conclusion}

\newpage
{
    \small
    \bibliographystyle{ieeenat_fullname}
    \bibliography{main}
}

%********** Supplementary ******************
\newpage
% \section{Technical Appendices and Supplementary Material}

\twocolumn[
\begin{center}
\LARGE \textbf{Technical Appendices and Supplementary Material} \\
\vspace{1em}
\end{center}
]

\input{sec/supplementary}

\end{document}

%% file: preamble.tex
\usepackage{booktabs}
\usepackage[table]{xcolor} 
\usepackage{array}
\newcolumntype{P}[1]{>{\raggedright\arraybackslash}p{#1}}

%% file: sec/0_abstract.tex
\begin{abstract}
Appearance-based gait recognition has achieved strong performance on controlled datasets, yet systematic evaluation of its robustness to real-world corruptions and silhouette variability remains lacking. We present RobustGait, a framework for fine-grained robustness evaluation of appearance-based gait recognition systems. RobustGait evaluation spans four dimensions: the type of perturbation (digital, environmental, temporal, occlusion), the silhouette extraction method (segmentation and parsing networks), the architectural capacities of gait recognition models, and various deployment scenarios. The benchmark introduces 15 corruption types at 5 severity levels across CASIA-B, CCPG, and SUSTech1K, with in-the-wild validation on MEVID, and evaluates six state-of-the-art gait systems. We came across several exciting insights. First, applying noise at the RGB level better reflects real-world degradation, and reveals how distortions propagate through silhouette extraction to the downstream gait recognition systems. Second, gait accuracy is highly sensitive to silhouette extractor biases, revealing an overlooked source of benchmark bias. Third, robustness is dependent on both the type of perturbation and the architectural design. Finally, we explore robustness-enhancing strategies, showing that noise-aware training and knowledge distillation improve performance and move toward deployment-ready systems.
% \footnote{Code and project page: \url{https://github.com/Reeshoon/RobustGait}}
\end{abstract}

%% file: sec/1_intro.tex
\section{Introduction}
\label{sec:intro}

\begin{figure}[t]
    \centering
    \includegraphics[width=\linewidth]{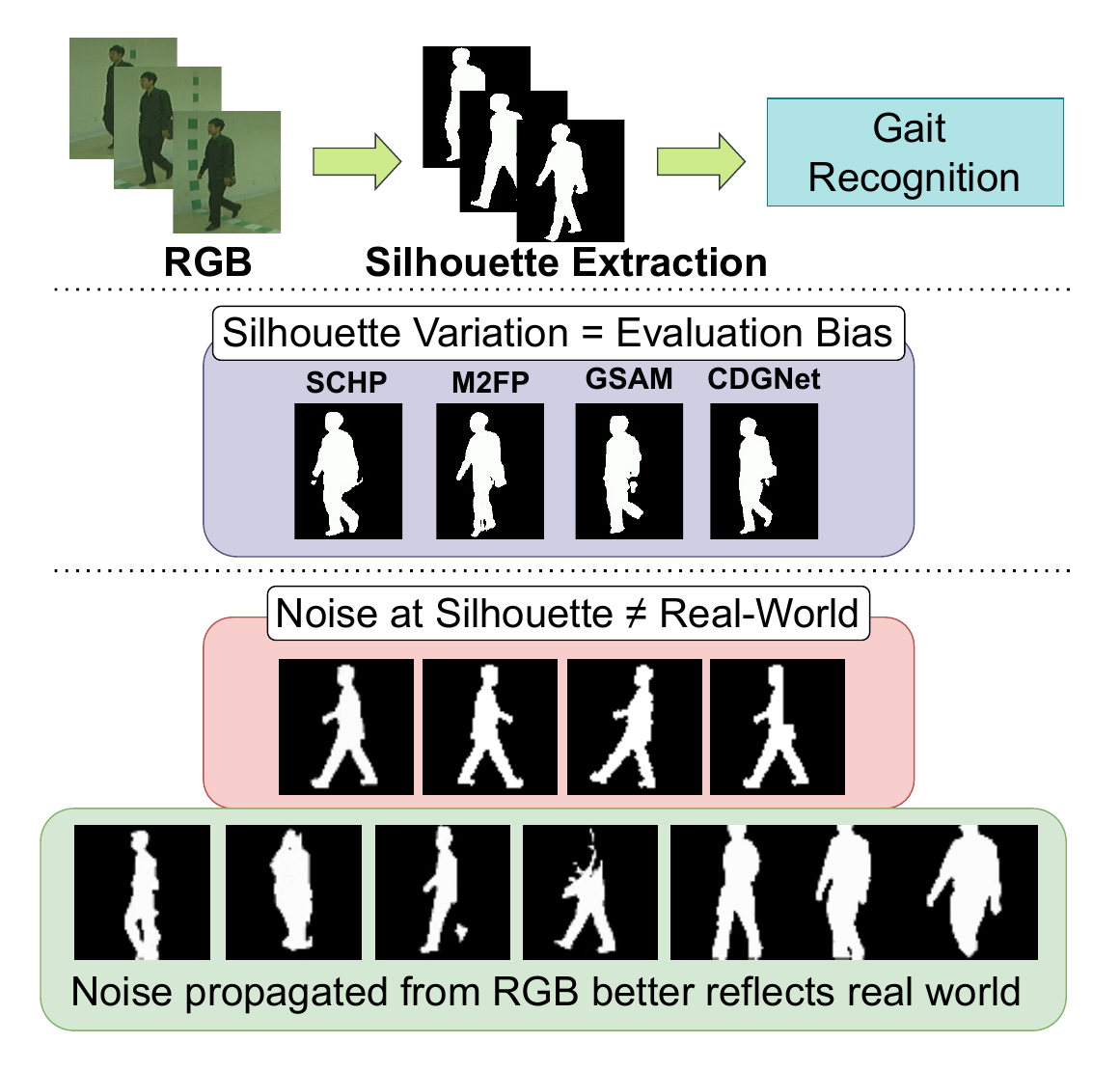}
    \caption{\textbf{Overlooked biases in appearance-based gait benchmarks.} 
    (i) Variation across silhouette extractors leads to evaluation bias due to variable silhouette quality, motivating the need for standardized extraction across benchmarks.
    (ii) directly applying noise to silhouettes restricts corruptions to simple augmentations as flipping, rotation, or erasing, whereas injecting noise at the RGB level allows various temporal, environmental, and digital degradations to propagate to silhouettes, better reflecting real-world scenarios.}

    \label{fig:teaser}
\end{figure}

Gait recognition aims to identify individuals based on their unique walking patterns captured from video sequences. Unlike face, fingerprint, or iris, gait can be captured at long range and is difficult to conceal, making it highly suitable for security, surveillance, and forensic applications. Despite strong progress of gait recognition models on datasets collected under controlled laboratory conditions \cite{casiab,oumvlp,ccpg}, wide-scale deployment in real-world scenarios remains limited\cite{grew}. Unconstrained videos are affected by a wide range of distortions \cite{grew,gait3d,zheng2023parsing,zeng2024benchmarking}, including camera noise~\cite{noise_impact}, lighting and weather variations~\cite{env_impact}, temporal inconsistencies across frames, and occlusions caused by objects obstructing the subject~\cite{occ_impact}. These factors create significant distribution shift between training benchmarks and deployment environments. Yet existing gait recognition models \cite{opengait,deepgait} are seldom evaluated under such circumstances, leaving a critical gap in understanding their real-world robustness.

Within gait recognition, existing methods can be broadly categorized into two groups: appearance-based approaches~\cite{gaitgl,deepgait}, which rely on human silhouette-based representations, and model-based approaches~\cite{fan2024skeletongait,zhang2023spatial}, which use 2D/3D human poses or SMPL-based reconstructions\cite{Zheng_2022_CVPR}. While both approaches have advanced the field, appearance-based methods remain the dominant choice in practice~\cite{fan2025opengait}. They are particularly effective in real-world scenarios, as they can operate reliably on low-resolution videos, avoid the need for accurate pose estimation, and are generally more computationally efficient~\cite{Li_2020_ACCV,Li2024GaitMFR,TemporalGaitSurvey}. Therefore, in our work, we focus on appearance-based gait.

The key challenges in comprehensively evaluating robustness of gait systems arise from gait recognition being a two step process \cite{opengait,fan2025opengait}, where the intermediate silhouette representation has to be extracted from RGB data as shown in Fig.\ref{fig:teaser}. First, noise in RGB videos propagate through the silhouette extraction stage and directly affect the quality of the silhouette representations. Hence, unlike standard computer vision tasks where robustness is often analyzed by perturbing the input image directly\cite{ImageNetC,Hendrycks2021ICCV,Tong_2021_CVPR}, gait recognition presents a unique challenge due to its dependence on intermediate representations. Directly applying naive augmentations to silhouette data, such as random erasing\cite{Wang2024QAGait}, fails to capture the complex perturbations introduced in the RGB input~\cite{opengait}.

Second, gait datasets vary widely in their silhouette generation pipelines. Older datasets such as CASIA-B~\cite{casiab} and OU-MVLP~\cite{oumvlp} rely on outdated background subtraction methods~\cite{casiab_segment}, whereas recent datasets like CCPG~\cite{ccpg} and SUSTech1K~\cite{sustech} employ modern segmentation architectures such as U-Net~\cite{unet} and PaddleSeg~\cite{paddleseg}. These discrepancies highlight the evolving nature of silhouette extraction and highlight a potential source of bias in gait recognition benchmarks. 

Recent benchmarks, such as GREW \cite{grew} and Gait3D \cite{gait3d}, have introduced in-the-wild datasets to advance gait recognition evaluation beyond controlled laboratory settings. However, a systematic evaluation of gait recognition models' robustness to various noise types, silhouette extractors, and model architectures is lacking. To address this, we present a comprehensive robustness analysis, evaluating the impact of silhouette extraction models on silhouette quality and gait recognition performance. Silhouette extractors are applied in a zero-shot manner without task-specific adaptation. Controlled noise augmentations are applied to RGB inputs before silhouette extraction to simulate realistic degradations that affect the extraction process. RobustGait spans four dimensions of evaluation: (i) perturbation type (digital, environmental, temporal, occlusion), (ii) silhouette extraction method (four representative segmentation and parsing networks), (iii) recognition architecture (sequence-based CNNs, set-based CNNs, transformers), and (iv) deployment scenarios (cross-extractor and cross-scene).

In summary, our main contributions are:  
\begin{itemize}  
\item We present \textbf{RobustGait}, a comprehensive benchmark spanning three widely used gait datasets: CASIA-B~\cite{casiab}, CCPG~\cite{ccpg}, and SUSTech1K~\cite{sustech}, under 15 corruption types and 5 severity levels.  
\item We simulate realistic degradations by introducing controlled corruptions at the RGB level, allowing noise to propagate naturally through the silhouette extraction stage.  
\item We systematically analyze the role of silhouette extraction, showing how differences across \textbf{four} extraction models introduce evaluation bias and affect recognition robustness. 
\item We evaluate \textbf{six} state-of-the-art gait recognition models across diverse architectures, revealing how robustness varies with corruption type, severity, and choice of extractor.  
\item We investigate robustness-improving strategies, including noise-aware training and knowledge distillation, and highlight their effectiveness as well as the trade-offs between robustness and clean-data accuracy.  
\end{itemize}

%% file: sec/2_related_work.tex
\section{Related Work}
\label{sec:related_work}
\begin{figure*}[t]
    \centering
    \includegraphics[width=\linewidth, keepaspectratio]{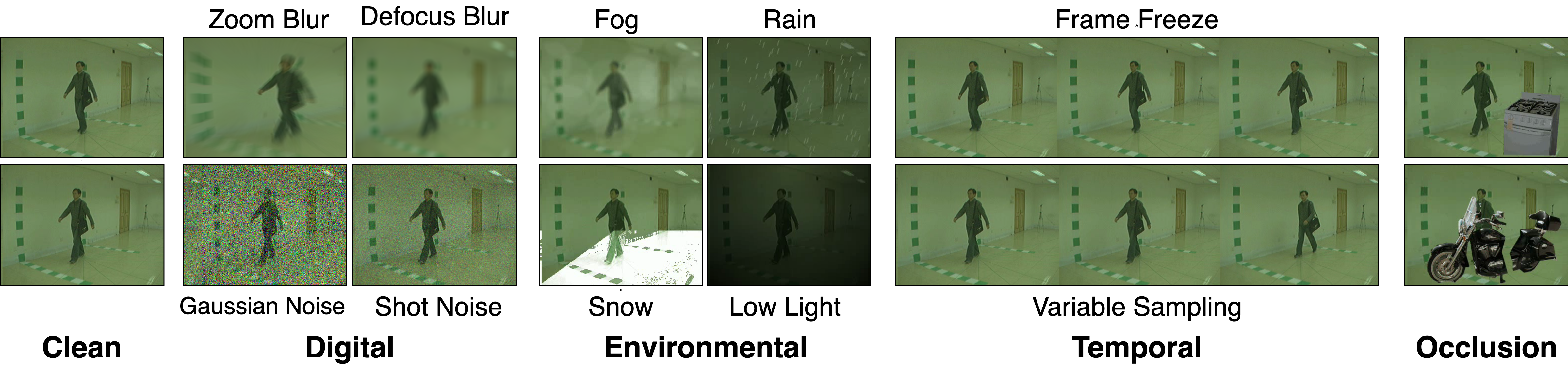}
    \caption{\textbf{Overview of noises}: Qualitative visualization of four major taxonomy of noises studied in our benchmark.}
    \label{fig:overview}
\end{figure*}
\noindent\textbf{Silhouette Extraction Models:}  
extract silhouette from RGB data essential for gait recognition. Early gait datasets like CASIA-B~\cite{casiab} generated silhouettes using background subtraction~\cite{casiab_segment}, which requires heuristic post-processing. Later improvements, such as CASIA-B*\cite{gaitedge}, aimed to refine alignment but still rely on outdated techniques. Recent datasets such as CCPG\cite{ccpg} and SUSTECH-1k~\cite{sustech} adopt the segmentation models U-Net~\cite{unet} and PaddleSeg~\cite{paddleseg}, while in-the-wild datasets GREW~\cite{grew} and Gait3D~\cite{gait3d} employ HTC~\cite{htc} and HRNet~\cite{hrnet} to achieve more robust silhouette extraction under challenging conditions.
\textbf{\textit{End-to-end gait recognition}} frameworks~\cite{biggait, gaitedge} attempt to learn intermediate representations from RGB data within the network. While these systems show promise, they often struggle to disentangle gait-relevant features from appearance noise of RGB data(e.g., clothing texture, lighting), leading to reduced robustness in cross-domain scenarios~\cite{gaitedge}. Recent studies~\cite{zheng2023parsing,gait3d} highlight that \textbf{\textit{human parsing models}}, designed to segment fine-grained body parts, lead to better robustness in the wild. Parsing methods fall into two categories: single human parsing (SHP)~\cite{e2ehumanparse,sslhumanparse,multilabelparse}, which processes one subject per frame using attention or pose guidance, and multiple human parsing (MHP)~\cite{contextualmhp,mhpres}, which handles multi-person scenarios with bottom-up reasoning. \textbf{\textit{Unlike prior works}}, that evaluate gait models using datasets from different parsing sources, our study ensures a \textit{fair comparison} by maintaining consistent silhouette quality across all models.

\noindent\textbf{Biometrics Robustness Benchmarks:}  In videos, there has been existing works \cite{kumar2022end, Dave_2022_WACV, modi2022video, Singh_Rana_Kumar_Vyas_Rawat_2024, kumar2025stable, rana2025omvid, kumar2025contextual, garg2025stpro} to understand videos at fine-grained level. Recently, robustness analysis becomes crucial as deep learning models transition to real-world applications in videos \cite{Kumar_2025_CVPR}, especially in biometrics. Face recognition studies establish comprehensive benchmarks through frameworks like FACESEC~\cite{Tong_2021_CVPR}, which evaluates various perturbation types and attack scenarios, while~\cite{Perez_2023_CVPR} examines semantic robustness via latent manipulations. Video-based tasks address real-world data shifts~\cite{vidrob1} and occlusion effects~\cite{vidrob2} in action recognition. In gait recognition, existing work evaluates only on isolated factors like clothing~\cite{gaitrob1}, viewing angles~\cite{gaitrob2}, and occlusions~\cite{gaitrob4}, but \textit{lacks analysis of compound effects} during silhouette extraction. Our work \textbf{\textit{differentiates itself}} by evaluating gait robustness across multiple types of real-world RGB noise, tracing how such perturbations propagate through the silhouette-extraction stage to impact final recognition accuracy. 

%% file: sec/3_benchmark_setup.tex
\section{RobustGait Benchmark}
\label{sec:benchmark_setup}

In this section, we discuss the details of our benchmark. Sec.~\ref{sec:perturb} describe the perturbations used in our study. Sec.~\ref{sec:bench_details} discuss the details of the datasets, network architectures, and evaluation metrics.

\begin{figure*}
    \centering
    \includegraphics[width= \linewidth]{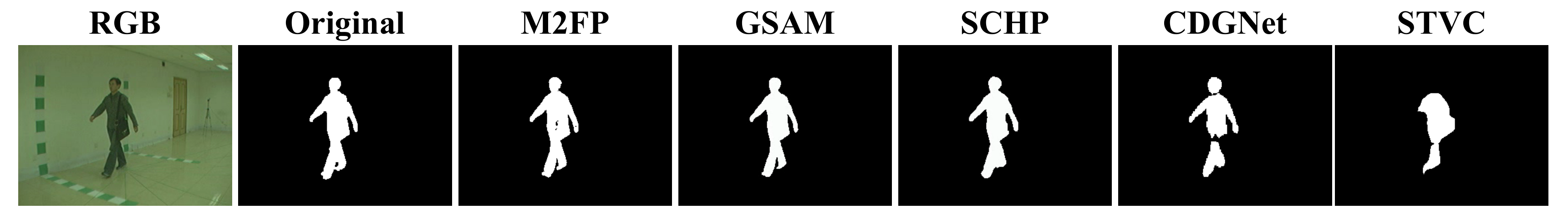}
    \caption{Qualitative analysis of parsing models on CASIA-B. Silhouette quality decreases from left to right. M2FP, SCHP, and GSAM
preserve body structure, while CDGNet and STVC show degradation.
    }
    \label{fig:segmod}
\end{figure*}

\subsection{Gait Corruption in Videos}
\label{sec:perturb}

Recent studies have highlighted diverse video degradations as major obstacles for gait recognition in real-world surveillance \cite{Zhao2024SituDiv,Li2024OcclusionSurvey,Mughal2025Covariates}.
Motivated by these findings, we examine four key noise categories: digital corruptions (camera artifacts and compression errors), environmental perturbations (lighting changes and weather effects), temporal distortions (frame-rate fluctuations and motion jitter), and occlusions (partial or full view blockage). An overview of perturbations is shown in Fig.~\ref{fig:overview}. \textbf{Digital} corruptions simulate \textit{sensor-induced artifacts}, encompassing various blurs (zoom, defocus, motion) and noise patterns (Gaussian, shot, impulse, speckle) which are widely modeled in corruption benchmarks such as Mini-Kinetics-C \cite{Wang2021ImageNetCVideo} and SSV2-C \cite{Xiao2022SSV2C}. \textbf{Environmental} perturbations emulate adverse conditions such as low light, fog, rain, and snow, consistent with prior studies \cite{Zhao2024SituDiv,Paul2013Surveillance} highlighting \textit{weather and illumination} as major failure factors in surveillance and person re-identification.
\textbf{Temporal} corruptions affect frame consistency through freezing, variable sampling rates, and focal zoom, reflecting common recording anomalies that \textit{degrade sequential models} \cite{Sun2024TemporalNoise,Liu2024OccludedReID}.
\textbf{Occlusion} corruptions introduce static foreground objects that \textit{partially obstruct} the subject, a challenge frequently studied in both gait and person re-identification contexts \cite{Li2024OcclusionSurvey,Liu2024OccludedReID}. These perturbations are implemented across five severity levels, ranging from level~I (least severe) to level~V (most severe). Detailed severity specifications and qualitative analysis are provided in the supplementary. 
% Fig. illustrates the comprehensive set of perturbations used in our analysis.

\subsection{Benchmark Details}
\label{sec:bench_details}

\begin{table}[!t]
  \centering
  \scriptsize
  \setlength{\tabcolsep}{3pt}
  \renewcommand{\arraystretch}{0.9}
  \begin{tabular}{r cccccc cc cc}
    \toprule
    \rowcolor{mygray}
    \textbf{Dataset} & \textbf{\#Ids} & \textbf{\#Seq.} & \textbf{\#Cam.} &
    \textbf{RGB} & \textbf{\#Cov.} &
    \multicolumn{2}{c}{\textbf{Env.}} &
    \multicolumn{2}{c}{\textbf{Setup}} \\
    \rowcolor{mygray} & & & & & & \textbf{Ctrl.} & \textbf{Wild} & \textbf{In} & \textbf{Out} \\
    \midrule
    CASIA\text{-}B~\cite{casiab}   & 124   & 13.6k & 11  & \checkmark & 2 & \checkmark &  & \checkmark &  \\
    SUSTech1K~\cite{sustech}       & 1050  & 25.2k & 12  & \checkmark & 7 & \checkmark &  &  & \checkmark \\
    CCPG~\cite{ccpg}               & 200   & 16.0k & 10  & \checkmark & 2 & \checkmark &  & \checkmark & \checkmark  \\
    MEVID~\cite{ccpg}               & 158   & 8.1k & 33  & \checkmark & - &  & \checkmark &  & \checkmark  \\
    \bottomrule
  \end{tabular}
  \caption{\textbf{Datasets Stats:}
  \#Ids, \#Seq., \#Cam. and \#Cov. denote number of identities, sequences, cameras and covariates. Environment (Env.) setup is divided between controlled (Ctrl.) or in-the-wild (wild). Gait setup is captured across indoor (In), outdoor (Out) or both. $^{\dagger}$ denotes unconstrained environment.}
  \label{tab:dataset_comparison}
\end{table}

\noindent \textbf{\textit{Datasets:}} Table~\ref{tab:dataset_comparison} provides detailed summary of the popular gait datasets. For fair evaluation of gait recognition models under simulated noise, we utilize datasets that provide raw RGB videos.  With newer datasets becoming increasingly available with RGB data, our datasets cover a wide range of environmental conditions and real-world scenarios. CASIA-B~\cite{casiab} represents a controlled indoor setup, CCPG~\cite{ccpg} is collected in a hybrid indoor-outdoor environment and SUSTech1K~\cite{sustech} offers an outdoor setting. These datasets capture varied covariates across different settings,e.g. clothing, carrying, umbrella, uniform, enabling a comprehensive evaluation of robustness when combined with real-world noise. To simulate realistic degradations for our study, we construct perturbed variants of CASIA-B, CCPG, and SUSTech1K, covering all possible scenarios for gait captures in varied conditions applying 15 types of corruptions at five severity levels (from mild to extreme). For in-the-wild evaluation, we use the MEVID~\cite{mevid} dataset, which is a large-scale surveillance dataset capturing subjects across diverse outdoor environments and unconstrained camera views.

\noindent \textit{\textbf{Architectures:}} Gait recognition proceeds in two stages: \textit{\textbf{silhouette extraction}} and \textit{\textbf{person re-identification}}. 
Guided by recent findings that human parsing improves human segmentation accuracy~\cite{CCGR,zheng2023parsing}, we evaluate both conventional segmentation models and dedicated human-parsing approaches for silhouette extraction. 
For segmentation, we adopt \textbf{Grounded SAM}~\cite{sam,groundedsam}, selected for its strong and consistent performance across diverse tasks~\cite{sam_gud}. 
For human parsing, drawing on the comprehensive survey in~\cite{human_parsing_survey}, we select the top models in two categories: Single-Human Parsing (SHP): \textbf{SCHP}~\cite{schp} and \textbf{CDGNet}~\cite{cdgnet}; and Multiple-Human Parsing (MHP): \textbf{M2FP}~\cite{m2fp}. We exclude \textbf{STVC} \cite{stvc}, due to its comparatively poor performance. For the \textit{gait recognition} stage, we build on the OpenGait repository~\cite{gaitbase}, which aggregates recent state-of-the-art methods. 
We benchmark six appearance-based models spanning different architectural paradigms and capacities. Among CNN-based networks, we include small-capacity models: \textbf{GaitPart}~\cite{Fan2020GaitPartTP} (1.2M parameters), \textbf{GaitGL}~\cite{gaitgl} (3.3M), and \textbf{GaitSet}~\cite{Chao2018GaitSetRG} (2.6M), as well as medium-capacity models: \textbf{GaitBase}~\cite{gaitbase} (7.4M) and \textbf{DeepGaitV2}~\cite{deepgait}(8.4M).To cover transformer architectures, we add the high-capacity model \textbf{SwinGait}~\cite{deepgait}(11M).

\noindent \textbf{\textit{Evaluation Metrics:}} We evaluate our benchmark on two metrics, namely:  \textbf{ID Retrieval:} Following query-gallery setup from prior works \cite{evaluate,opengait}, we report Rank-1 retrieval accuracy. It measures the proportion of probe samples whose highest-scoring match in the reference set shares the correct identity, and, \textbf{Robustness metric:} to evaluate the robustness of models against perturbations. Given, performance of models on clean ($\mathcal{D}_c$) and perturbed dataset ($\mathcal{D}_p$), we calculate absolute robustness as $\delta_a = 1 - \tfrac{\mathcal{D}_c - \mathcal{D}_p}{100}$ and relative robustness as $\delta_r = 1 - \tfrac{\mathcal{D}_c - \mathcal{D}_p}{\mathcal{D}_c}$. Absolute robustness ($\delta_a$) is the total percentage drop in performance, while relative robustness ($\delta_r$) is the proportional drop compared to the clean baseline. \textbf{IoU recognition}:We quantify silhouette quality as the Intersection-over-Union between the segmentation-generated mask and the original mask.

%% file: sec/4_benchmarking_analysis.tex
\section{Benchmark Analysis}
\label{sec:benchmarking_analysis}

\begin{figure}[t!]
  \centering
  \includegraphics[width= \linewidth]{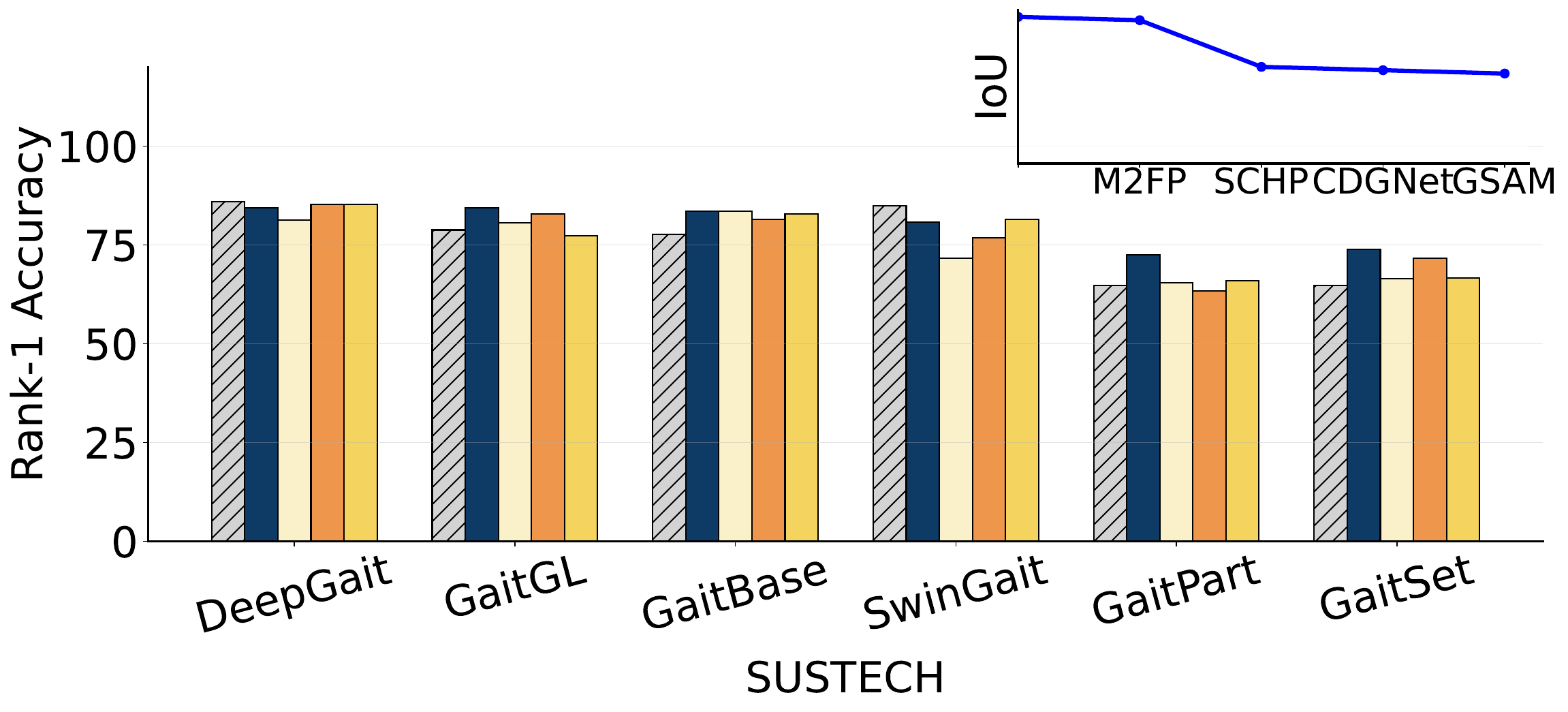}
  \caption{Impact of silhouette segmentation on gait recognition on SUSTECH. }
  \label{fig:segmentation2}
\end{figure}

\begin{figure*}[t!]
    \centering
    \includegraphics[width= \linewidth]{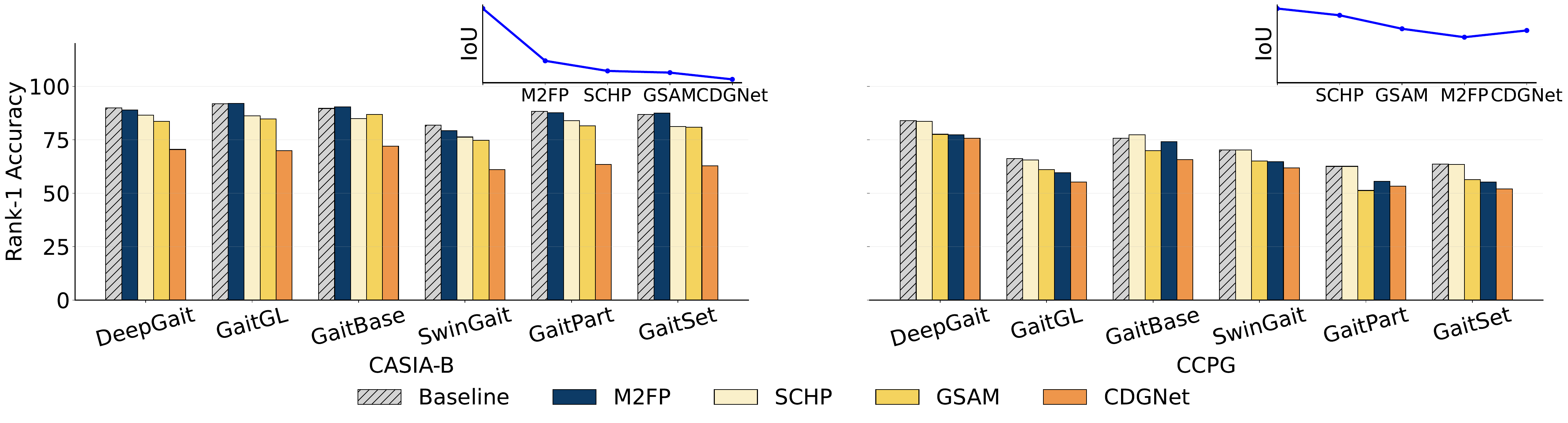}
    \caption{Impact of silhouette segmentation on gait recognition. Left (CASIA-B and Right (CCPG). The \textcolor{blue}{IoU curve} positively correlates with recognition performance: segmentation methods with higher IoU generally yield higher Rank-1 accuracy across gait models (e.g., high-IoU M2FP on CASIA-B; SCHP on CCPG), while lower-IoU methods (e.g., CDGNet) correspond to lower accuracies. This highlights that better silhouette masks improve downstream gait recognition on those silhouettes.}

    \label{fig:segmentation1}
\end{figure*}

% Legend identical to Fig.~\ref{fig:segmentation1} 
\subsection{Impact of Silhouette Extraction}

\noindent \textbf{Motivation:} 
Silhouette extraction critically determines gait recognition accuracy, yet existing datasets employ heterogeneous and often outdated extraction pipelines (e.g., background subtraction or handcrafted segmentation \cite{unet, paddleseg}). 
Evaluating recognition models on such disparate inputs obscures true architectural performance and risks misleading conclusions about datasets themselves. 
We therefore study modern segmentation and human-parsing extractors to assess their impact on recognition, aiming for fair, low-intervention evaluation without heavy post-processing. Fig.\ref{fig:segmod} shows how silhouette quality varies across different segmentation models.

\begin{figure*}[t!]
    \centering
    \includegraphics[width=\linewidth]{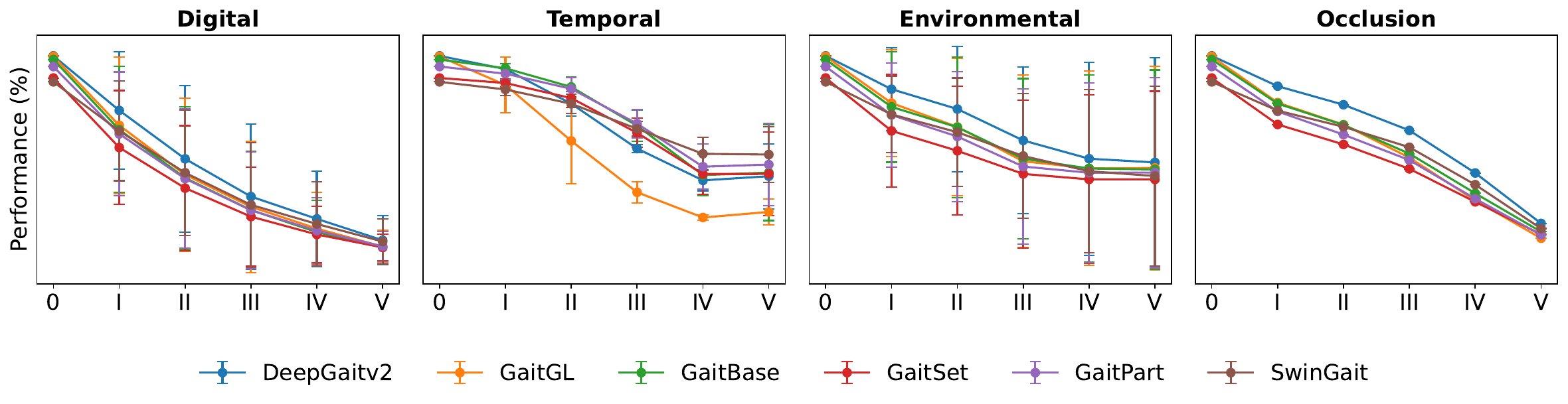}
    \caption{\textbf{Impact of Noise severity} (CASIA-B): Performance degrades with increase in noise severity. Models are \textit{most} robust to environment to environmental changes and \textit{least} to digital modification.}
    \label{fig:severity_comp}
\end{figure*}

\noindent \textbf{Observations:} Figures  \ref{fig:segmentation2} and \ref{fig:segmentation1} compares four parsing models (described in Sec.~\ref{sec:benchmark_setup}) across three benchmark datasets: CASIA-B, CCPG, and SUSTech. We achieve following observations: (1) \textbf{\textit{Different silhouette extractors leads to unfair comparison}}.  The figures shows that Gait models are sensitive to silhouette quality with significant variations in performance. It proves our hypothesis that maintaining same silhouette extractor is necessary. (2) \textbf{\textit{Extractor choice drives recognition performance}}. Original silhouettes do not consistently yield the highest accuracy. On CASIA-B and SUSTech1K, \textbf{M2FP} surpasses the baseline across most conditions, while \textbf{SCHP} leads on CCPG. Silhouette quality, measured by Intersection-over-Union (IoU), mirrors these trends: M2FP achieves the highest IoU on CASIA-B and SUSTech1K, and SCHP attains the best IoU on CCPG.

\subsection{Robustness against Noisy Dataset}
\label{sec:gait_downstream}

\begin{figure*}[t!]
    \centering
    \includegraphics[width=\linewidth]{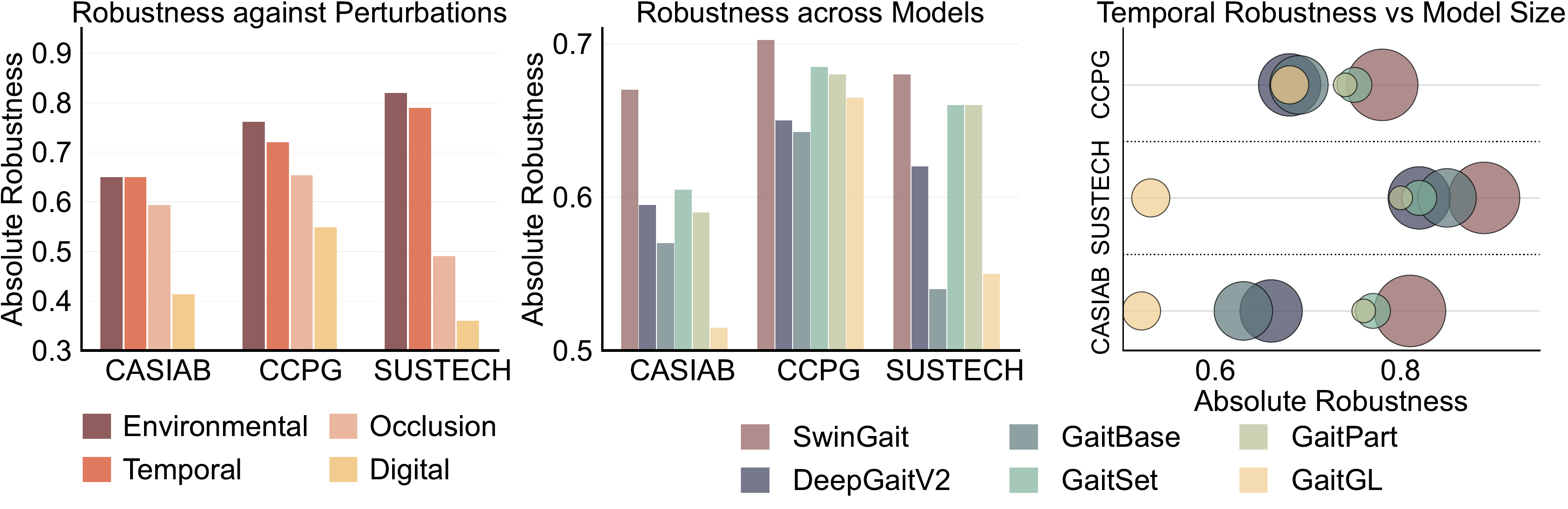}
    \caption{\textbf{Robustness across Model Architecture}
    \textit{(Left)} Models are most robust against environmental and temporal corruptions in general.
    \textit{(Mid)} SwinGait is the most resilient model amongst all. \textit{(Right)} Smaller capacity set-based models are more robust than the larger capacity models. The size of plotted points is proportional to the capacity of the model. The plot shows temporal robustness across models. 
    }
    \label{fig:robustness_analysis}
\end{figure*}

\noindent \textbf{Motivation:} Real-world videos contain temporal artifacts, environmental perturbations, occlusions, and digital distortions, all of which degrade silhouettes and hinder gait recognition~\cite{Zhao2024SituDiv,Mughal2025Covariates}. 
Models trained on controlled datasets often fail when deployed in such unconstrained settings. To expose these vulnerabilities, we perturb only the probe set while keeping the gallery clean, isolating the effect of real-world noise on recognition accuracy. Silhouettes are extracted with \textbf{SCHP}~\cite{schp}, selected for its strong accuracy efficiency balance (see supplementary for details). Our findings are as follows:

\noindent \textbf{\textit{Local Distortions are Most Damaging:}} Digital corruptions (e.g., blur, compression) and occlusion consistently cause the sharpest performance decline (Fig.~\ref{fig:severity_comp}). As their severity increases, it leads to heavily dispersed feature clusters (Supplementary Fig. 10), breaking discriminative boundaries and degrading identity separability. This vulnerability is further amplified when gallery and probe sets come from mismatched distributions (Fig.~\ref{fig:robustness_analysis}, left). Together, these findings reveal a fundamental weakness: current gait features are highly brittle to pixel-level corruption and distribution shifts, limiting their robustness in uncontrolled surveillance scenarios.

\noindent \textbf{\textit{Sequential and Structural Cues Provide Natural Robustness:}} In contrast, perturbations that preserve structural integrity, such as temporal noise or environmental effects (fog, rain, or snow) are less harmful (Fig. \ref{fig:severity_comp}). Temporal corruptions disrupts only subsets of frames within the whole sequence, leaving sequential redundancy that models can exploit to recover missing cues. Environmental noise primarily alters global visibility through low-frequency changes but leaves body contours intact, enabling gait models to rely on motion dynamics rather than fine-grained appearance. Consequently, these conditions, while realistic, impose only modest penalties compared to local distortions.

\noindent \textbf{\textit{Mismatch between clean gallery and noisy probe exposes hidden fragility.}}  
Figure~\ref{fig:robustness_analysis} (left) shows that gait models are especially vulnerable to digital noise and occlusion, yet remain comparatively stable under environmental or temporal perturbations. This performance gap highlights how distribution shifts clean gallery versus corrupted probe can sharply degrade recognition, underscoring the need for feature representations that remain reliable when training and deployment conditions differ.

\subsection{Robustness across Model Architecture}
\label{label:model_arc}

\noindent \textbf{Motivation:} Robustness depends not only on noise type but also on the underlying architectural design \cite{Tong_2021_CVPR} in biometrics, which governs how gait cues are captured across spatial and temporal dimensions. Clean-benchmark accuracy alone can mask weaknesses, as different architectures vary in their ability to model local continuity, temporal dynamics, and global context. By evaluating robustness across diverse models, we identify which designs generalize best and clarify the trade-offs that inform the development of more reliable gait recognition systems.

\noindent \textbf{\textit{Higher clean accuracy does not guarantee robustness}}. Figure~\ref{fig:robustness_analysis} (mid) shows that although several CNN-based models achieve strong clean accuracy, SwinGait consistently delivers higher absolute robustness across all datasets. Patterns in Sec.~\ref{sec:gait_downstream} explain why Transformer-based models are more robust under noise (Fig.~\ref{fig:robustness_analysis}, mid). Local distortions break CNNs, while Transformers use global self-attention for compensation. SwinGait’s hybrid design CNNs for local features, Transformers for spatiotemporal context ensures consistent robustness across datasets.

\noindent \textbf{\textit{Temporal modeling drives robustness to sequence noise.}} Sequence noise introduces variations in frame sampling. Figure~\ref{fig:robustness_analysis} (right) shows that set-based models like \textbf{GaitSet} remain stable under frame sampling and freezing because they treat gait as an unordered frame set, avoiding fragile frame-to-frame dependencies.  Sequence-based CNNs, in contrast, suffer large drops as their reliance on local temporal order breaks under missing or corrupted frames extra capacity only deepens this vulnerability.  Hybrid Transformers such as \textbf{SwinGait} combine CNN local feature extraction with global self-attention across time, letting uncorrupted frames compensate for disruptions and preserving discriminative cues, which explains their consistent robustness across datasets.

\begin{figure}[t]
  \centering
  \includegraphics[width= 0.8\linewidth]{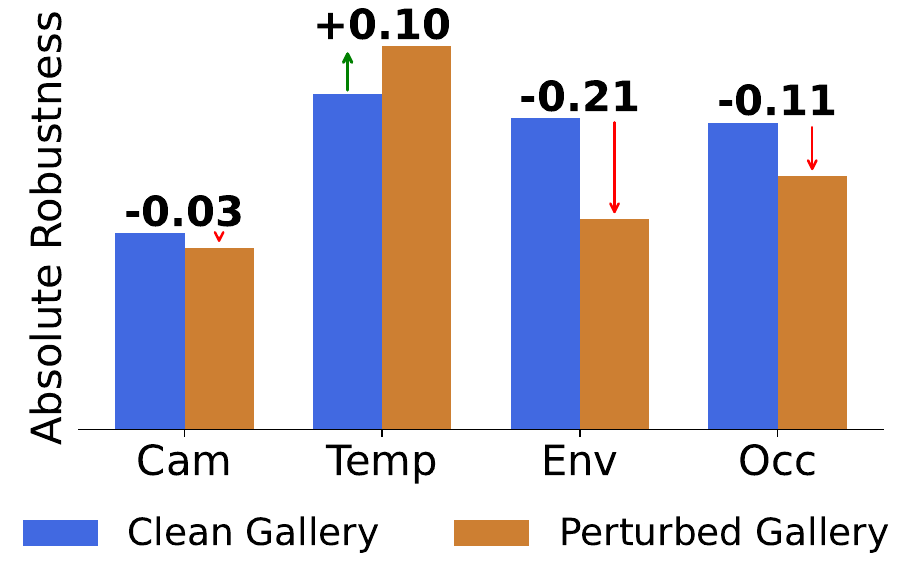}
  \caption{\textbf{Absolute robustness} analysis with \textcolor{blue}{clean} vs. \textcolor{brown}{noisy} gallery on CASIA-B.}
  \label{fig:noisy_gal}
\end{figure}

\begin{figure*}[t!]
    \centering
    \includegraphics[width=\linewidth]{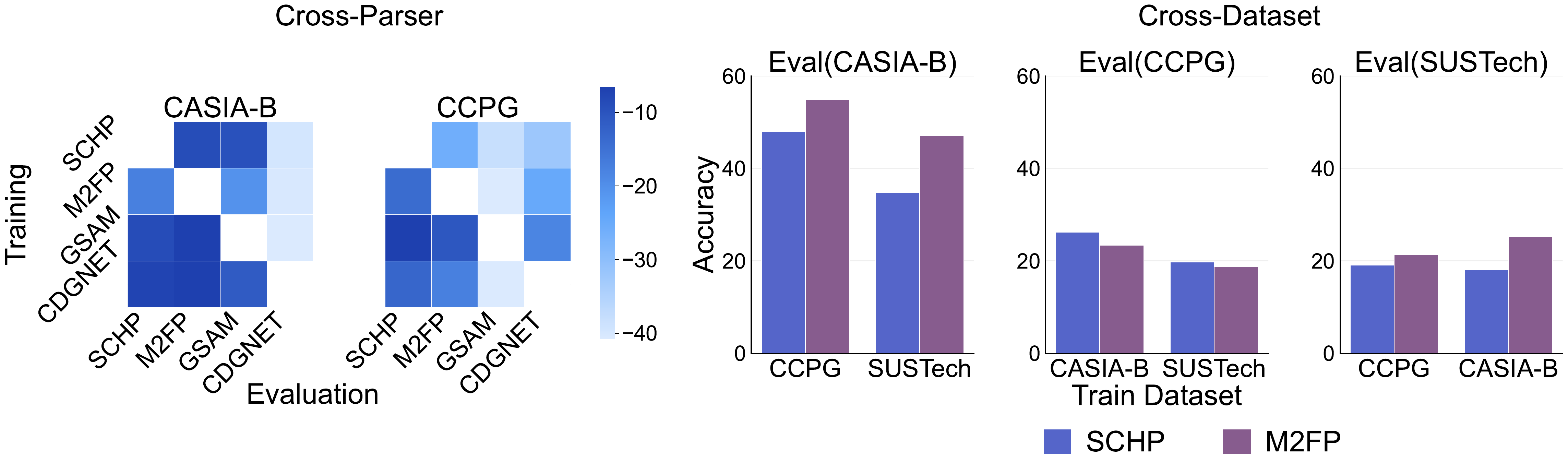}
    \caption{\textbf{Cross Parsing Evaluation} on DeepGaitv2. \textit{(Left)} Accuracy heatmap showing performance drops when training and evaluation use different parsing models on the same dataset, indicating strong parser dependency. \textit{(Right)} Parser effectiveness when evaluated in a cross dataset setup. M2FP performs better on CASIA-B and SUSTech, while SCHP excels on CCPG. }
    \label{fig:cross_parsing}
\end{figure*}

\label{sec:gait_upstream}

\subsection{Deployment Scenarios}
\textbf{Motivation:}In practical deployment scenarios, silhouette extraction models may differ across environments due to variations in segmentation backbones, pre-processing setups, or domain conditions\cite{opengait,zhang2022realgait}. Recent large-scale person re-identification and surveillance datasets \cite{mevid,Briar_2023_WACV} highlight these challenges by capturing subjects in unconstrained, cross-camera surveillance settings. \\

\noindent \textbf{Impact of noise in gallery data:}  
In this scenario, both probe and gallery data is affected by noise, occlusions, or environmental variation. This pertains to situations such as outdoor cameras and surveillance videos that suffers from turbulence or environmental noise. To evaluate the robustness of models when gallery is noisy, we create a fixed gallery consisting of different noises with different severities. We evaluate the robustness of the models by using each of the noisy dataset as the query against this fixed dataset. 
Fig. \ref{fig:noisy_gal} shows that both the absolute and relative robustness degrades for digital, environmental, and occlusion perturbations. Thus, we infer that \textbf{\textit{models trained on clean data overfit to clean features.}} When the model is primarily trained on clean data, it struggles to handle noisy gallery and noisy probe scenarios. A clean gallery acts as a stabilizer, allowing the model to extract better features from the noisy probe. Noise in gallery data degrade features, making gait matching  challenging.

\noindent \textbf{Cross-Silhouette-Extraction Evaluation:}  
In this setting, a gait model trained on silhouettes generated by one extraction method is tested on silhouettes produced by a different method. The results (Fig.~\ref{fig:cross_parsing}) reveal a clear drop in accuracy when train and test silhouettes come from different extraction pipelines, demonstrating that gait models perform poorly under such mismatches. This highlights the strong dependence of gait recognition models on the silhouette extraction method and their limited generalization across shifts.

\noindent \textbf{Cross-Dataset Evaluation:}  
Here, the same silhouette extraction method is used, but the training and evaluation datasets differ. Results in Fig.~\ref{fig:cross_parsing} (right) show that SCHP consistently yields the best performance on CCPG, while M2FP achieves higher accuracy on CASIA-B and SUSTech1K. This indicates that silhouette extraction effectiveness is mutually dependent on dataset structure, and dataset-specific characteristics strongly influence which extraction method is most suitable.

%% file: sec/5_improving_robustness.tex
% \section{Defending Against Silhouette Corruptions}
\section{Analysing augmentation and distillation for robustness}
\label{sec:solution_distill}

To improve the robustness of gait recognition models under silhouette corruptions, we propose and evaluate two strategies: noise-aware training and an efficient adaptation using student-teacher distillation. We further validate their effectiveness on a large-scale MEViD\cite{mevid} dataset which has real-world corruptions, demonstrating scalability to real-world settings.

% Robustness Across Augmentation Ratios
\begin{figure*}[t]
    \centering
    \includegraphics[width=0.95\linewidth]{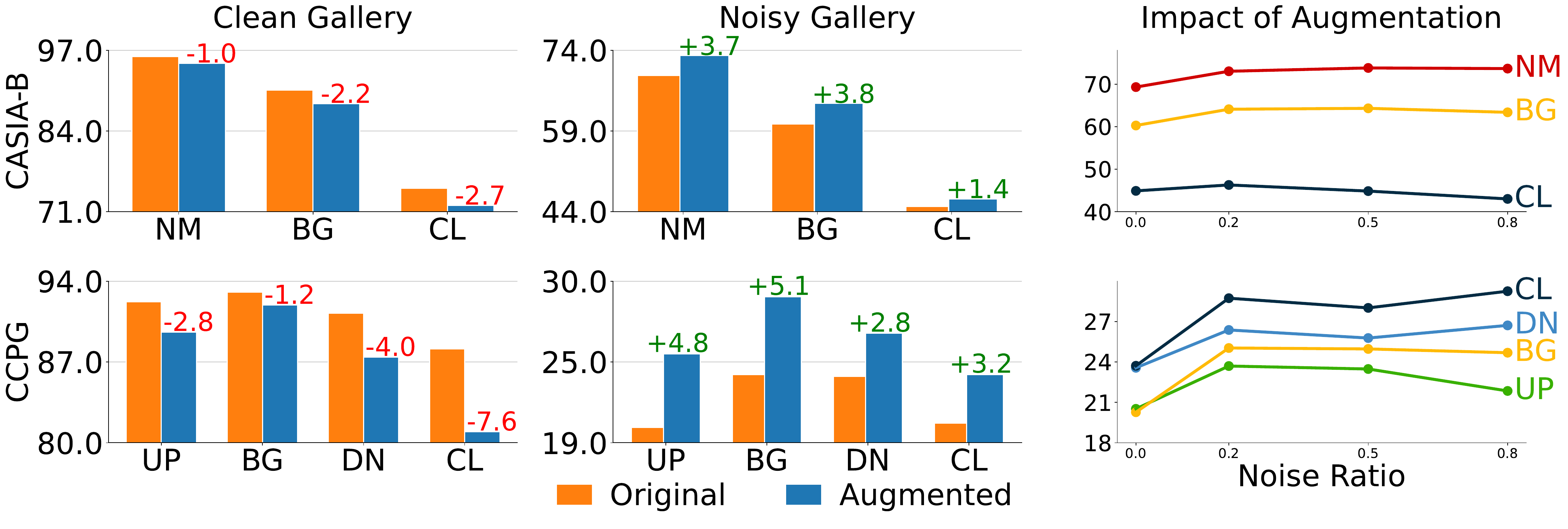}
    \caption{\textbf{Noise Aware Training.}
    \textit{Top row}: CASIA-B trained with a mix of clean and noisy data. 
    \textit{Bottom row}: CCPG dataset under identical settings.  
    Left: Accuracy when gallery is clean. 
    Middle: Accuracy when gallery is noisy.
    Right: Accuracy trends with different augmentation ratios.}
    \label{fig:casia_ccpg_augmentation}
\end{figure*}

\subsection{Noise Aware Training}

We study the effect of adding noisy silhouettes, derived from perturbed RGB inputs, into the training process. This differs from conventional augmentations such as flips or random erasing, which are directly applied to silhouettes. We see the following observations:  (1) \textbf{\textit{Training with noisy data improves robustness but induces forgetting}}. Fig.~\ref{fig:casia_ccpg_augmentation} show that models trained on a mix of clean and noisy data become more resilient to perturbations than those trained solely on clean inputs. However, this robustness comes with a slight loss of accuracy on clean test sets, indicating that exposure to noise can cause the model to partially forget clean-domain representations. (2) \textbf{\textit{Efficient training achieved with limited noisy data}}. 
Fig.~\ref{fig:casia_ccpg_augmentation} shows that introducing only a small fraction of noisy samples during training provides nearly the same robustness gains as extensive noise augmentation. Performance improvements plateau around 25-30\% noisy data, indicating diminishing returns and highlighting the \textit{efficiency of this limited-noise strategy}.

\begin{figure}[t]
    \centering
    \includegraphics[width= \columnwidth]{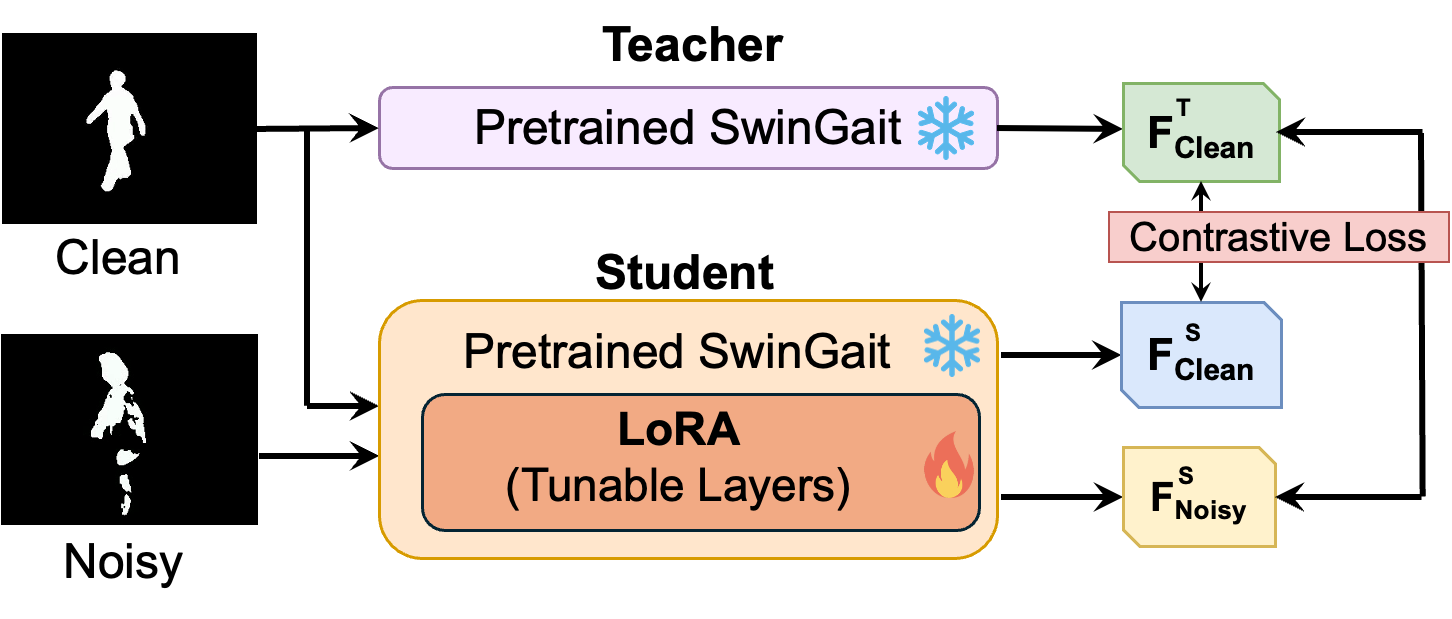}
    \caption{\textbf{Distillation framework} for robustness: The student with LoRA layers learns to align noisy embeddings with the clean-teacher representations.}
    \label{fig:solution}
\end{figure}

\subsection{Efficient distillation for noise-robust Gait}

We further explore the use of distillation to improve resilience to silhouette noise while preserving clean-data performance. We propose a knowledge distillation framework that adapts a model using Low-Rank Adaptation (LoRA) \citep{hu2022lora}. Fig.~\ref{fig:solution} shows our framework adapts \textbf{SwinGait}, chosen for its strong robustness. A frozen teacher processes clean silhouettes to produce stable feature embeddings, while a student with identical backbone trains only its LoRA modules. The student receives both clean and noisy silhouettes and optimizes a dual loss: a CLIP-style contrastive loss on clean inputs to match the teacher, and a consistency loss aligning noisy-student embeddings with clean-teacher embeddings. This setup promotes noise-invariant representations without sacrificing clean performance. Results in Table~\ref{tab:casiab_side} show that the distilled model preserves clean test accuracy comparable to a clean-trained model while achieving superior performance under noise. This indicates that distillation enhances robustness without the accuracy loss typically caused by training solely on noisy data, effectively \textbf{\textit{mitigating the forgetting}} observed in direct noisy training.

\noindent
\textbf{Scaling to large-scale real-world dataset:}
We extend our study to the large-scale Multi-view Extended Videos with Identities (MEVID) dataset \citep{mevid}  to demonstrate the practical effectiveness of our benchmark beyond synthetic datasets. MEVID is a challenging video person re-identification (ReID) dataset designed for real-world settings, incorporating diverse indoor and outdoor environments, multiple camera viewpoints, and significant clothing variations. We compare our training strategies in a zero-shot setting, where models trained on CASIA-B are directly evaluated on MEVID. As shown in Table~\ref{tab:mevid_side}, training on noisy data yields higher Top-5 accuracy (13.7\%) than the clean baseline (11.1\%), indicating improved generalization. Training using the proposed distillation framework contribute further gains across metrics, with 7.5 mAP, 9.2 Top-1, and 18.1 Top-5 accuracy. These results suggest that \textbf{\textit{robustness techniques developed on synthetic datasets can transfer effectively to real-world conditions}} when paired with appropriate training strategies.

\begin{table}[t]
    \centering
    \renewcommand{\arraystretch}{1.}
    \footnotesize
    \begin{tabular}{r|ccc|ccc}
        \specialrule{1.5pt}{0pt}{0pt}
        \rowcolor{gray!20}
        \textbf{Train Method} & \multicolumn{3}{c|}{\textbf{Test on Clean}} & \multicolumn{3}{c}{\textbf{Test on Noisy}} \\
        \rowcolor{gray!10}
        & NM & BG  & CL  & NM  & BG  & CL  \\
        \hline
        Baseline       & \textbf{90.9} & \textbf{82.6} & \textbf{62.1} & 66.4 & 55.3 & \underline{38.3} \\
        Noise Aware          & 85.9          & 73.2          & 46.1 & \textbf{76.3} & \textbf{63.4} & \textbf{42.1} \\
        Distillation   & \underline{89.6} & \underline{79.9} & \underline{57.0} & \underline{71.3} & \underline{59.1} & \underline{38.3} \\
        \specialrule{1.5pt}{0pt}{0pt}
    \end{tabular}\
    \caption{\textbf{Performance of SwinGait}  on CASIA-B under different conditions and training setups.}
    \label{tab:casiab_side}
\end{table}

\begin{table}[t]
    \centering
    \renewcommand{\arraystretch}{1.}
    \footnotesize
    \begin{tabular}{r|ccc}
        \specialrule{1.5pt}{0pt}{0pt}
        \rowcolor{gray!20}
        \textbf{Train Method} & \textbf{mAP} & \textbf{Top-1} & \textbf{Top-5} \\
        \hline
        Baseline       & 5.4          & 6.3            & 11.1 \\
        %Noisy          & 6.7 & \underline{7.0} & 12.1 \\
        Noise Aware  & \underline{7.0}          & \textbf{9.2}    & \underline{13.7} \\
        Distillation   & \textbf{7.5} & \textbf{9.2}    & \textbf{18.1} \\
        \specialrule{1.5pt}{0pt}{0pt}
    \end{tabular}
    \caption{\textbf{Zero-shot generalization} to MEVID.}
    \label{tab:mevid_side}
\end{table}

%% file: sec/6_conclusion.tex
\section{Conclusion}
\label{sec:conclusion}
In this study, we systematically analyze the robustness of gait recognition models to RGB noise, focusing on both key components: parsing models and gait models. Our benchmark establishes a standardized framework for dataset selection and model evaluation across parsing models and performance of gait models in real-world scenarios, ensuring fair comparisons. Through extensive analysis, we provide valuable insights into the impact of parsing models and the resilience of gait models under real-world perturbations. We believe this study will serve as a foundation for future research, advancing the understanding of robustness in gait recognition.

%% file: sec/supplementary.tex
We provide additional results, implementation details, and expanded tables for various experiments discussed in the main paper. It includes results across different parsing models, perturbation robustness analysis, and further implementation specifics.
\begin{itemize}
    \item \textbf{Section~\ref{sec:discussion}}: Discussion on ethical issues and broader impact of the work.
    \item \textbf{Section~\ref{sec:extra_details}}: Technical details on benchmark setup including segmentation models, training configuration, and evaluation protocol.
    \item \textbf{Section~\ref{sec:limitations}}: Limitations of our study.
    \item \textbf{Section~\ref{sec:broader_impact}}: Discussion on societal and ethical implications of our study.
    \item \textbf{Section~\ref{sec:add_results}}: Additional results across parsing models and robustness scores on multiple severity levels and noisy training.
    \item \textbf{Section~\ref{sec:distillation}}: Training details of proposed approach.
    \item \textbf{Section~\ref{sec:mevid}}: Gait model evaluation and training results on the large-scale MEVID benchmark.
    \item \textbf{Section~\ref{sec:details}}: Implementation specifics and severity schedules for the 15 corruption types in our benchmark.
\end{itemize}

\section{Discussion}
\label{sec:discussion}
\textbf{Ethical issues:} The datasets used in this study: CASIA-B, CCPG, SUSTech1K, and MEVID, are publicly available benchmark datasets, each of which has addressed ethical considerations in their respective publications. Our use of these datasets strictly follows their intended research purposes.
\textbf{Broader impact:} By investigating robustness in real-world surveillance scenarios, this work highlights both opportunities and risks. While the proposed techniques improve the reliability of gait recognition systems, they may also accelerate broader deployment in contexts where privacy, consent, and ethical considerations are critical. As with any biometric technology, careful evaluation of use cases and their societal implications is essential to ensure responsible adoption.

\section{Benchmark Details}
\label{sec:extra_details}
\noindent \textbf{Segmentation Setup:}
We use five parsing models for silhouette extraction. SCHP and M2FP both use a ResNet101 backbone; M2FP is initialized with CIHP-pretrained weights. CDGNet also uses a ResNet101 backbone, initialized with LIP-pretrained weights. STVC is configured with a ResNet18 backbone and stride-8 settings. GSAM is included as a foundation segmentation model. All outputs are converted to binary silhouettes for downstream use.

\noindent \textbf{Gait Recognition Setup:}
We adopt the OpenGait framework and default hyperparameters to train six models: GaitBase, GaitGL, DeepGait, SwinGait, GaitPart, and GaitSet. Input silhouettes are resized to 64×44 with 30-frame sequences (10 for DeepGait on SUSTech1K). Training uses $P \times K$ sampling, standard triplet + cross-entropy loss, and optimizers chosen per model: SGD for DeepGait/GaitBase/GaitSet, Adam for GaitGL/GaitPart, and weighted Adam for SwinGait. Frame skipping is applied as per the original model configurations.

\noindent \textbf{Implementation Details:}
All experiments were conducted using 2 Tesla V100-PCIE-32GB GPUs (CUDA 12.4). Our implementation is based on the OpenGait repository:
\url{https://github.com/ShiqiYu/OpenGait}

\noindent \textbf{Evaluation Protocol:} Gait recognition models are evaluated using a standard probe-gallery setup, where the gallery set contains reference sequences for each identity, and the probe set contains query sequences to be matched against the gallery. The goal is to correctly identify each probe sequence by retrieving the most similar sequence from the gallery. For the \textbf{CASIA-B} dataset, the gallery includes sequences captured under the normal walking condition: \texttt{nm-01}, \texttt{nm-02}, \texttt{nm-03}, and \texttt{nm-04}. The probe set comprises sequences from varying conditions, including \texttt{nm-05}, \texttt{nm-06}, \texttt{bg-01}, \texttt{bg-02}, \texttt{cl-01}, and \texttt{cl-02}. In the \textbf{CCPG} dataset, the gallery consists of sequences \texttt{U1\_D1}, \texttt{U2\_D2}, \texttt{U3\_D3}, \texttt{U0\_D3}, and \texttt{U0\_D0}. The probe set includes \texttt{U0\_D0\_BG}, \texttt{U0\_D0}, \texttt{U3\_D3}, \texttt{U1\_D0}, and another instance of \texttt{U0\_D0\_BG} under different conditions. For the \textbf{SUSTech1K} dataset, the gallery contains the sequence \texttt{00-nm}, while the probe set includes a diverse set of variations: \texttt{01-nm}, \texttt{bg}, \texttt{cl}, \texttt{cr}, \texttt{ub}, \texttt{uf}, \texttt{oc}, \texttt{nt}, and additional sequences labeled \texttt{01} through \texttt{04}.

\section{Limitations}
\label{sec:limitations}
Due to differences in generalization, not all parsing models were applicable across all three datasets (CASIA-B, CCPG, and SUSTech1K), leading to selective parser usage in certain evaluations. The study focuses on silhouette-based gait recognition, leaving out pose and depth-based methods. At higher noise severities, the data becomes heavily degraded, making meaningful evaluation on those data challenging. Additionally, the computational cost of evaluating multiple parsing and recognition models across datasets and perturbation settings constrained the depth and exhaustiveness of some analyses presented in this work.

\section{Broader Impact}
\label{sec:broader_impact}
This work highlights the need for robust gait recognition systems for practical deployment by systematically evaluating model performance under real-world corruptions such as occlusions, lighting changes, and sensor noise. It demonstrates that parsing and recognition models vary significantly in robustness, encouraging more transparent benchmarking. The released dataset support future research in robust and generalizable biometric representation learning.

While the proposed techniques improve robustness, they may also facilitate broader deployment of gait recognition systems in real-world settings, including those where ethical, privacy, or consent considerations are important. As with any biometric technology, thoughtful evaluation of use cases and societal implications is essential to ensure responsible use.

\section{Additional Results}
\label{sec:add_results}

\textbf{Results on Different Silhouette Extraction Models:} Silhouette quality plays an important role in the performance of gait recognition models. We evaluate the impact of different human parsing models on gait recognition across three datasets: CASIA-B, CCPG, and SUSTech1K. Results for each combination of parsing method and gait recognition model are shown in Tables~\ref{tab:casiab_parsing}, \ref{tab:ccpg_parsing}, and \ref{tab:sustech_parsing}.
\begin{table*}[t]
\centering
\begin{adjustbox}{max width=\textwidth}
\begin{tabular}{l|ccc|ccc|ccc|ccc|ccc}
\rowcolor{mygray}
\hline
 & \multicolumn{3}{c|}{\textbf{Baseline}\textcolor{lightgray}{\scriptsize{[ICPR06]}}} 
 & \multicolumn{3}{c|}{\textbf{SCHP}\textcolor{lightgray}{\scriptsize{[TPAMI20]}}} 
 & \multicolumn{3}{c|}{\textbf{M2FP}\textcolor{lightgray}{\scriptsize{[arXiv23]}}} 
 & \multicolumn{3}{c|}{\textbf{CDGNet}\textcolor{lightgray}{\scriptsize{[CVPR22]}}} 
 & \multicolumn{3}{c}{\textbf{GSAM}\textcolor{lightgray}{\scriptsize{[arXiv24]}}} \\
\rowcolor{mygray}
\textbf{IoU} 
 & \multicolumn{3}{c|}{1.00} 
 & \multicolumn{3}{c|}{0.63} 
 & \multicolumn{3}{c|}{0.69} 
 & \multicolumn{3}{c|}{0.58} 
 & \multicolumn{3}{c}{0.62} \\
\hline
Method 
 & NM & BG & CL 
 & NM & BG & CL 
 & NM & BG & CL 
 & NM & BG & CL 
 & NM & BG & CL \\
\hline
DeepGaitV2 \textcolor{lightgray}{\scriptsize{[arXiv23]}} & 97.3 & 93.8 & \textbf{78.5} & 95.0 & 89.8 & 74.6 & \textbf{97.8} & \textbf{93.9} & 75.4 & 88.8 & 74.6 & 48.2 & 94.0 & 87.3 & 69.5 \\
GaitGL\textcolor{lightgray}{\scriptsize{[ICCV21]}}   & 97.4 & 94.5 & \textbf{83.6} & 93.2 & 88.1 & 77.4 & \textbf{97.6} & \textbf{94.7} & \textbf{83.6} & 83.5 & 73.3 & 53.0 & 92.6 & 86.6 & 75.1 \\
GaitBase\textcolor{lightgray}{\scriptsize{[CVPR23]}} & 97.6 & 94.0 & 77.4 & 93.6 & 88.4 & 72.8 & \textbf{98.1} & \textbf{95.2} & \textbf{77.9} & 89.3 & 76.7 & 49.9 & 96.0 & 89.9 & 74.7 \\
SwinGait\textcolor{lightgray}{\scriptsize{[arXiv23]}} &   94.0   &    87.1  &  64.4    &  88.8    &  80.8    &  59.2    &    93.7  &   85.9   &   58.3   &  82.7    &   64.9   &   35.7   &  89.6    &   79.9   &   54.8   \\
GaitPart\textcolor{lightgray}{\scriptsize{[CVPR20]}} & \textbf{96.2} & \textbf{90.6} & 78.2 & 92.5 & 85.9 & 73.3 & 96.0 & \textbf{90.6} & \textbf{76.6} & 79.4 & 65.9 & 45.2 & 91.9 & 81.8 & 70.8 \\
GaitSet\textcolor{lightgray}{\scriptsize{[AAAI19]}}  & 95.6 & 90.2 & 74.8 & 91.7 & 83.2 & 68.6 & \textbf{96.5} & \textbf{90.8} & \textbf{75.2} & 80.9 & 65.6 & 41.8 & 92.9 & 80.8 & 68.9 \\
\hline
\end{tabular}
\end{adjustbox}
\caption{Comparison of six gait recognition models under different silhouette extraction methods on CASIA-B. Results are reported under three conditions: NM (normal), BG (bag), and CL (clothing). The best result per row and condition is highlighted in bold.}
\label{tab:casiab_parsing}
\end{table*}

\begin{table*}[t]
\centering
\begin{adjustbox}{max width=\textwidth}
\begin{tabular}{l|cccc|cccc|cccc|cccc|cccc}
\rowcolor{mygray}
\hline
 & \multicolumn{4}{c|}{\textbf{Baseline}\textcolor{lightgray}{\scriptsize{[CVPR23]}}} 
 & \multicolumn{4}{c|}{\textbf{SCHP}\textcolor{lightgray}{\scriptsize{[TPAMI20]}}} 
 & \multicolumn{4}{c|}{\textbf{M2FP}\textcolor{lightgray}{\scriptsize{[arXiv23]}}} 
 & \multicolumn{4}{c|}{\textbf{CDGNet}\textcolor{lightgray}{\scriptsize{[CVPR22]}}} 
 & \multicolumn{4}{c}{\textbf{GSAM}\textcolor{lightgray}{\scriptsize{[arXiv24]}}} \\
\rowcolor{mygray}
\textbf{IoU} 
 & \multicolumn{4}{c|}{1.00} 
 & \multicolumn{4}{c|}{0.96} 
 & \multicolumn{4}{c|}{0.83} 
 & \multicolumn{4}{c|}{0.87} 
 & \multicolumn{4}{c}{0.88} \\
\hline
Method 
 & CL & UP & DN & BG 
 & CL & UP & DN & BG 
 & CL & UP & DN & BG 
 & CL & UP & DN & BG 
 & CL & UP & DN & BG \\
\hline
DeepGait\textcolor{lightgray}{\scriptsize{[arXiv23]}} & \textbf{79.2} & \textbf{85.0} & \textbf{81.3} & \textbf{90.0} & 78.9 & 84.8 & 81.0 & 89.7 & 70.8 & 76.6 & 76.0 & 85.8 & 67.5 & 75.8 & 74.7 & 84.7 & 71.3 & 78.5 & 77.0 & 83.6 \\
GaitGL\textcolor{lightgray}{\scriptsize{[ICCV21]}}   & \textbf{61.8} & \textbf{68.1} & \textbf{64.6} & \textbf{70.2} & 61.7 & 67.7 & 62.7 & 70.1 & 53.9 & 61.0 & 59.5 & 63.8 & 49.8 & 56.3 & 53.8 & 61.2 & 55.9 & 63.6 & 57.7 & 66.9 \\
GaitBase\textcolor{lightgray}{\scriptsize{[CVPR23]}} &72.1 & 75.3 & 77.2 & 78.7 & \textbf{73.8} & \textbf{76.9} & \textbf{78.1} & \textbf{80.7} & 70.8 & 73.1 & 76.9 & 75.9 & 60.6 & 64.8 & 68.4 & 69.3 & 64.2 & 69.6 & 72.6 & 73.1 \\
SwinGait\textcolor{lightgray}{\scriptsize{[arXiv23]}} & \textbf{61.2} & \textbf{71.9} & \textbf{66.5} & 81.5 & \textbf{61.2} & 71.4 & 66.0 & \textbf{82.1} & 54.0 & 63.4 & 64.6 & 77.0 & 49.5 & 61.8 & 60.1 & 75.9 & 55.0 & 65.4 & 64.1 & 76.0 \\
GaitPart\textcolor{lightgray}{\scriptsize{[CVPR20]}} & 57.7 & 63.6 & \textbf{62.8} & 66.4 &
\textbf{58.1} & \textbf{63.7} & 61.7 & \textbf{66.7} &
51.3 & 56.9 & 55.6 & 58.6 &
47.8 & 55.2 & 53.0 & 57.0 &
46.1 & 52.7 & 50.7 & 55.7 \\
GaitSet\textcolor{lightgray}{\scriptsize{[AAAI19]}}  & \textbf{58.8} & \textbf{64.5} & \textbf{63.7} & \textbf{67.5} & 59.2 & 63.7 & \textbf{63.9} & \textbf{67.1} & 50.8 & 55.8 & 56.7 & 58.0 & 46.4 & 52.7 & 53.8 & 55.1 &   51.5   &   57.7   &   57.0   &  59.2    \\
\hline
\end{tabular}
\end{adjustbox}
\caption{Comparison of six gait recognition models under different silhouette extraction methods on CCPG. Results are reported under four conditions: CL (full), UP (up), DN (down), and BG (bag). The best result per row and condition is highlighted in bold.}
\label{tab:ccpg_parsing}
\end{table*}

\begin{table*}[t]
\centering
\begin{adjustbox}{max width=\textwidth}
\begin{tabular}{l|cccc|cccc|cccc|cccc|cccc}
\rowcolor{mygray}
\hline
 & \multicolumn{4}{c|}{\textbf{Baseline}\textcolor{lightgray}{\scriptsize{[CVPR23]}}} 
 & \multicolumn{4}{c|}{\textbf{SCHP}\textcolor{lightgray}{\scriptsize{[TPAMI20]}}} 
 & \multicolumn{4}{c|}{\textbf{M2FP}\textcolor{lightgray}{\scriptsize{[arXiv23]}}} 
 & \multicolumn{4}{c|}{\textbf{CDGNet}\textcolor{lightgray}{\scriptsize{[CVPR22]}}} 
 & \multicolumn{4}{c}{\textbf{GSAM}\textcolor{lightgray}{\scriptsize{[arXiv24]}}} \\
\rowcolor{mygray}
\textbf{IoU} 
 & \multicolumn{4}{c|}{1.00} 
 & \multicolumn{4}{c|}{0.85} 
 & \multicolumn{4}{c|}{0.99} 
 & \multicolumn{4}{c|}{0.84} 
 & \multicolumn{4}{c}{0.83} \\
\hline
Method 
 & NM & CL & UM & OVR 
 & NM & CL & UM & OVR 
 & NM & CL & UM & OVR 
 & NM & CL & UM & OVR 
 & NM & CL & UM & OVR \\
\hline
DeepGaitV2\textcolor{lightgray}{\scriptsize{[arXiv23]}}
 & 89.1 & 76.9 & 86.2 & 85.9 
 & 86.0 & 68.3 & 82.7 & 81.3 
 & 89.0 & 51.4 & 87.8 & 84.4 
 & 88.7 & 45.6 & 86.6 & 85.3 
 & 82.2 & 58.9 & 88.0 & 85.3 \\
GaitBase\textcolor{lightgray}{\scriptsize{[CVPR23]}}
 & 80.4 & 62.9 & 74.9 & 77.7
 & 88.1 & 77.3 & 83.7 & 83.6 
 & 88.7 & 56.4 & 84.7 & 83.6 
 & 86.4 & 49.6 & 80.2 & 81.5 
 & 81.2 & 58.2 & 83.9 & 82.8    \\
SwinGait\textcolor{lightgray}{\scriptsize{[arXiv23]}}
 & 88.1 & 80.6 & 84.4 & 84.9 
 & 79.7 & 53.7 & 70.8 & 71.6 
 & 86.3 & 46.3 & 82.7 & 80.8 
 & 81.6 & 37.8 & 77.5 & 76.8 
 & 80.6 & 49.7 & 83.9 & 81.5   \\
GaitGL\textcolor{lightgray}{\scriptsize{[ICCV21]}} 
 & 83.6 & 70.2 & 76.9 & 78.8 
 & 85.0 & 78.2 & 81.5 & 80.6 
 & 87.5 & 74.8 & 83.8 & 84.4 
 & 81.2 & 58.2 & 83.9 & 82.8   
 & 75.7 & 56.7 & 79.5 & 77.3  \\
GaitSet\textcolor{lightgray}{\scriptsize{[AAAI19]}}
 & 62.6 & 27.7 & 62.1 & 64.7 
 & 72.6 & 39.9 & 63.1 & 66.5 
 & 78.9 & 40.7 & 71.1 & 73.9 
 & 79.7 & 53.7 & 70.8 & 71.6 
 & 69.1 & 31.8 & 64.9 & 66.7  \\
GaitPart\textcolor{lightgray}{\scriptsize{[CVPR20]}}
 & 63.7 & 29.8 & 61.3 & 64.8 
 & 69.6 & 46.8 & 62.5 & 65.4 
 & 76.5 & 47.5 & 69.1 & 72.5 
 & 67.0 & 29.9 & 58.2 & 63.3 
 & 63.7 & 34.7 & 62.9 & 65.9 \\ 
\hline
\end{tabular}
\end{adjustbox}
\caption{Comparison of six gait recognition models under different silhouette extraction methods on SUSTech1K. Results are reported under NM (normal), CL (clothing), UM (umbrella), and OVR (overall).}
\label{tab:sustech_parsing}
\end{table*}

\textbf{Robustness Analysis of Gait Recogniton Models under noises:} To assess the robustness of modern gait recognition models, we evaluate performance across three benchmark datasets (CASIA-B, CCPG, SUSTech1K) under four broad perturbation categories. For each model and dataset, we compute both absolute robustness ($\delta_a$) and relative robustness ($\delta_r$) as defined in the main paper. Table~\ref{tab:casiab_robustness} reports results on the CASIA-B dataset. Table~\ref{tab:ccpg_robustness} and Table~\ref{tab:sustech_robust} show the corresponding results for CCPG and SUSTech1K, respectively.

\label{sec:robust}
\begin{table*}[t!]
    \centering
    \renewcommand{\arraystretch}{1.06}
    \scalebox{1}{
    \setlength{\tabcolsep}{5pt} 
    \begin{tabular}{r| cc cc cc cc cc}
        \specialrule{1.5pt}{0pt}{0pt}
        \rowcolor{mygray}
        \cellcolor{mygray} \textbf{CASIAB}\textcolor{lightgray}{\scriptsize{[ICPR06]}} & \multicolumn{2}{c}{\textbf{Camera}} & \multicolumn{2}{c}{\textbf{Temporal}} & \multicolumn{2}{c}{\textbf{Environmental}} & \multicolumn{2}{c}{\textbf{Occlusion}} \\
        \rowcolor{mygray} & $\delta_a$ & $\delta_r$ & $\delta_a$ & $\delta_r$& $\delta_a$ & $\delta_r$& $\delta_a$ & $\delta_r$ \\
        \hline\hline
        DeepGaitV2\textcolor{lightgray}{\scriptsize{[arXiv23]}} & 0.42 & \textbf{0.33} & 0.66 & 0.61 & 0.67 & \textbf{0.62} & 0.63 & \textbf{0.57} \\
        GaitGL\textcolor{lightgray}{\scriptsize{[ICCV21]}} & 0.38 & 0.28 & 0.52 & 0.44 & 0.62 & 0.56 & 0.54 & 0.46 \\
        GaitBase\textcolor{lightgray}{\scriptsize{[CVPR23]}} & 0.37 & 0.26 & 0.72 & 0.67 & 0.63 & 0.56 & 0.56 & 0.49 \\
        GaitSet\textcolor{lightgray}{\scriptsize{[AAAI19]}} & 0.42 & 0.25 & 0.77 & 0.70 & 0.64 & 0.53 & 0.59 & 0.46 \\
        GaitPart\textcolor{lightgray}{\scriptsize{[CVPR20]}} & 0.40 & 0.27 & 0.76 & 0.70 & 0.63 & 0.55 & 0.57 & 0.47 \\
        SwinGait\textcolor{lightgray}{\scriptsize{[arXiv23]}} & \textbf{0.49} & 0.32 & \textbf{0.81} & \textbf{0.75} & \textbf{0.71} & 0.61 & \textbf{0.67} & 0.56 \\
        % Average & 0.41 & 0.29 & 0.70 & 0.64 & 0.65 & 0.57 & 0.59 & 0.50 \\
        \specialrule{1.5pt}{0pt}{0pt}
    \end{tabular}}
    \caption{Absolute ($\delta_a$) and relative ($\delta_r$) robustness scores of gait models on the CASIA-B dataset across different perturbation types. Higher is better.}
    \label{tab:casiab_robustness}
\end{table*}

\begin{table*}[t!]
    \centering
    \renewcommand{\arraystretch}{1.06}
    \scalebox{1}{
    \setlength{\tabcolsep}{5pt} 
    \begin{tabular}{r| cc cc cc cc cc}
        \specialrule{1.5pt}{0pt}{0pt}
        \rowcolor{mygray}
        \cellcolor{mygray} \textbf{CCPG}\textcolor{lightgray}{\scriptsize{[CVPR23]}} & \multicolumn{2}{c}{\textbf{Camera}} & \multicolumn{2}{c}{\textbf{Temporal}} & \multicolumn{2}{c}{\textbf{Environmental}} & \multicolumn{2}{c}{\textbf{Occlusion}} \\
        \rowcolor{mygray} & $\delta_r$ & $\delta_a$ & $\delta_r$ & $\delta_a$& $\delta_r$ & $\delta_a$& $\delta_r$ & $\delta_a$ \\
        \hline\hline
        DeepGaitV2\textcolor{lightgray}{\scriptsize{[arXiv23]}} & 0.63 & 0.55 & 0.61 & 0.68 & 0.71 & 0.76 & \textbf{0.53} & 0.61 \\
        GaitGL\textcolor{lightgray}{\scriptsize{[ICCV21]}} & 0.70 & 0.55 & 0.52 & 0.68 & 0.64 & 0.76 & 0.50 & 0.67 \\
        GaitBase\textcolor{lightgray}{\scriptsize{[CVPR23]}} & 0.65 & 0.54 & 0.61 & 0.69 & 0.65 & 0.73 & 0.50 & 0.61 \\
        GaitSet\textcolor{lightgray}{\scriptsize{[AAAI19]}} & 0.71 & 0.55 & 0.61 & 0.75 & 0.63 & 0.76 & 0.49 & 0.68 \\
        GaitPart\textcolor{lightgray}{\scriptsize{[CVPR20]}} & \textbf{0.72} & 0.54 & 0.58 & 0.74 & 0.62 & 0.76 & 0.48 & \textbf{0.68} \\
        SwinGait\textcolor{lightgray}{\scriptsize{[arXiv23]}} & 0.69 & \textbf{0.56} & \textbf{0.69} & \textbf{0.78} & \textbf{0.72} & \textbf{0.80} & 0.52 & 0.67 \\
        % Average & 0.41 & 0.29 & 0.70 & 0.64 & 0.65 & 0.57 & 0.59 & 0.50 \\
        \specialrule{1.5pt}{0pt}{0pt}
    \end{tabular}}
    \caption{Absolute ($\delta_a$) and relative ($\delta_r$) robustness scores of gait models on the CCPG dataset across different perturbation types.}
    \label{tab:ccpg_robustness}
\end{table*}

\begin{table*}[t!]
    \centering
    \renewcommand{\arraystretch}{1.06}
    \scalebox{1}{
    \setlength{\tabcolsep}{4pt} 
    \begin{tabular}{r| cc cc cc cc cc}
        \specialrule{1.5pt}{0pt}{0pt}
        \rowcolor{mygray}
        \cellcolor{mygray} \textbf{SUSTech1k}\textcolor{lightgray}{\scriptsize{[CVPR23]}} & \multicolumn{2}{c}{\textbf{Camera}} & \multicolumn{2}{c}{\textbf{Temporal}} & \multicolumn{2}{c}{\textbf{Environmental}} & \multicolumn{2}{c}{\textbf{Occlusion}} \\
        \rowcolor{mygray} & $\delta_r$ & $\delta_a$ & $\delta_r$ & $\delta_a$& $\delta_r$ & $\delta_a$& $\delta_r$ & $\delta_a$ \\
        \hline\hline
        DeepGaitV2\textcolor{lightgray}{\scriptsize{[arXiv23]}} & 0.19 & 0.38 & 0.76 & 0.82 & 0.69 & 0.76 & \textbf{0.34} & 0.50 \\
        SwinGait\textcolor{lightgray}{\scriptsize{[arXiv23]}} & 0.19 & \textbf{0.47} & 0.83 & 0.89 & \textbf{0.74} & \textbf{0.83} & 0.28 & \textbf{0.53} \\
        GaitBase\textcolor{lightgray}{\scriptsize{[CVPR23]}} & 0.19 & 0.33 & \textbf{0.89} & \textbf{0.91} & 0.70 & 0.75 & 0.32 & 0.43 \\

        \specialrule{1.5pt}{0pt}{0pt}
    \end{tabular}}
    \caption{Absolute ($\delta_a$) and relative ($\delta_r$) robustness scores of gait models on the SUSTech1k dataset across different perturbation types.}
    \label{tab:sustech_robust}
\end{table*}

\textbf{Additional results for noises:} 
Tables \ref{tab:deepgaitcasia}, \ref{tab:gaitbasecasia}, \ref{tab:swingaitcasia}, 
\ref{tab:deepgaitccpg}, \ref{tab:gaitbaseccpg}, \ref{tab:swingaitccpg}, 
\ref{tab:dg_sg_sustech_side}, \ref{tab:gaitbasesustech} show performance results 
of gait models under the different noise types at each of the 5 severity levels.

\begin{table}[t!]
    \centering
    \setlength{\tabcolsep}{4pt} 
    \resizebox{\linewidth}{!}{%
    \begin{tabular}{l|cccccc}
        \specialrule{1.0pt}{0pt}{0pt}
        \rowcolor{mygray}
        \textbf{Perturbations} & \textbf{Clean} & \textbf{Sev 1} & \textbf{Sev 2} & \textbf{Sev 3} & \textbf{Sev 4} & \textbf{Sev 5} \\ 
        \hline\hline
        gaussian\_noise   & 86.5 & 72.5 & 35.0 & 2.8 & 2.4 & 0.0 \\ 
        defocus\_blur     & 86.5 & 66.5 & 32.7 & 2.7 & 1.5 & 1.8 \\      
        zoom\_blur        & 86.5 & 35.3 & 19.4 & 8.5 & 4.1 & 2.6 \\ 
        impulse\_noise    & 86.5 & 15.5 & 3.0  & 1.9 & 0.5 & 0.0 \\
        impulse\_noise2   & 86.5 & 64.8 & 15.0 & 4.3 & 2.5 & 1.7 \\ 
        speckle\_noise    & 86.5 & 81.9 & 75.7 & 63.1 & 44.0 & 25.2 \\ 
        shot\_noise       & 86.5 & 85.3 & 81.5 & 69.9 & 45.8 & 6.4 \\ 
        zoom\_in          & 86.5 & 83.5 & 77.4 & 57.5 & 33.8 & 23.7 \\ 
        \hline
        freeze            & 86.5 & 77.3 & 62.4 & 45.7 & 30.1 & 44.9 \\ 
        sampling          & 86.5 & 84.4 & 69.7 & 48.0 & 36.4 & 25.2 \\ 
        \hline
        low\_light        & 86.5 & 86.1 & 86.6 & 86.6 & 86.5 & 86.5 \\ 
        rain              & 86.5 & 83.4 & 74.0 & 25.2 & 5.5 & 2.3 \\ 
        snow              & 86.5 & 72.1 & 69.3 & 66.5 & 68.8 & 72.1 \\ 
        fog               & 86.5 & 47.3 & 25.1 & 23.1 & 9.3 & 2.8 \\ 
        \hline
        Static            & 86.5 & 73.5 & 65.6 & 54.6 & 36.4 & 14.9 \\ 
        \specialrule{1.5pt}{0pt}{0pt}
    \end{tabular}%
    }
    \caption{DeepGait robustness scores on CASIA-B dataset (rounded to one decimal place).}
    \label{tab:deepgaitcasia}
\end{table}

\begin{table}[t!]
    \centering
    \setlength{\tabcolsep}{4pt} 
    \resizebox{\linewidth}{!}{%
    \begin{tabular}{l|cccccc}
        \specialrule{1.5pt}{0pt}{0pt}
        \rowcolor{mygray}
        \textbf{Perturbations} & \textbf{Clean} & \textbf{Sev 1} & \textbf{Sev 2} & \textbf{Sev 3} & \textbf{Sev 4} & \textbf{Sev 5} \\ 
        \hline\hline
        gaussian\_noise & 84.9 & 62.3 & 22.9 & 1.8 & 1.7 & 0.0 \\
        defocus\_blur   & 84.9 & 54.6 & 17.9 & 2.2 & 1.4 & 1.7 \\
        zoom\_blur      & 84.9 & 20.6 & 11.1 & 5.3 & 3.2 & 2.7 \\
        impulse\_noise  & 84.9 & 10.0 & 1.8  & 2.1 & 0.0 & 0.0 \\
        impulse\_noise2 & 84.9 & 54.1 & 10.1 & 2.8 & 1.6 & 1.5 \\
        speckle\_noise  & 84.9 & 76.0 & 67.4 & 50.6 & 29.7 & 14.5 \\
        shot\_noise     & 84.9 & 82.0 & 75.3 & 59.5 & 31.8 & 3.8 \\
        zoom\_in        & 84.9 & 80.3 & 68.2 & 40.4 & 19.8 & 12.7 \\ 
        \hline
        freeze          & 84.9 & 79.8 & 70.3 & 51.9 & 29.4 & 51.1 \\ 
        sampling        & 84.9 & 82.5 & 76.1 & 61.4 & 41.7 & 22.1 \\ 
        \hline
        low\_light      & 84.9 & 84.6 & 84.9 & 84.9 & 85.0 & 85.0 \\
        rain            & 84.9 & 78.4 & 65.2 & 17.0 & 4.9 & 1.5 \\ 
        snow            & 84.9 & 64.2 & 59.9 & 55.6 & 58.4 & 62.8 \\ 
        fog             & 84.9 & 31.6 & 13.8 & 12.6 & 5.1 & 2.3 \\ 
        \hline
        Static          & 84.9 & 66.1 & 57.0 & 44.7 & 27.7 & 11.3 \\ 
        \specialrule{1.5pt}{0pt}{0pt}
    \end{tabular}%
    }
    \caption{GaitBase robustness scores on CASIA-B dataset (rounded to one decimal place).}
    \label{tab:gaitbasecasia}
\end{table}

\begin{table}[t!]
    \centering
    \setlength{\tabcolsep}{4pt} 
    \resizebox{\linewidth}{!}{%
    \begin{tabular}{l|cccccc}
        \specialrule{1.5pt}{0pt}{0pt}
        \rowcolor{mygray}
        \textbf{Perturbations} & \textbf{Clean} & \textbf{Sev 1} & \textbf{Sev 2} & \textbf{Sev 3} & \textbf{Sev 4} & \textbf{Sev 5} \\ 
        \hline\hline
        gaussian\_noise & 75.4 & 61.6 & 30.9 & 3.0 & 2.2 & 0.0 \\ 
        defocus\_blur   & 75.4 & 55.0 & 25.7 & 2.3 & 1.5 & 1.9 \\ 
        zoom\_blur      & 75.4 & 29.3 & 16.0 & 6.6 & 3.6 & 2.4 \\ 
        impulse\_noise  & 75.4 & 15.2 & 2.7  & 2.0 & 0.4 & 0.0 \\ 
        impulse\_noise2 & 75.4 & 57.8 & 15.0 & 4.1 & 1.5 & 1.2 \\ 
        speckle\_noise  & 75.4 & 70.5 & 65.7 & 54.6 & 39.2 & 23.9 \\ 
        shot\_noise     & 75.4 & 73.7 & 70.3 & 60.5 & 40.1 & 7.2 \\         
        zoom\_in        & 75.4 & 72.0 & 66.1 & 49.3 & 28.6 & 20.5 \\ 
        \hline
        freeze          & 75.4 & 70.2 & 63.1 & 53.0 & 39.6 & 52.8 \\ 
        sampling        & 75.4 & 74.0 & 68.9 & 57.6 & 49.6 & 35.8 \\ 
        \hline
        low\_light      & 75.4 & 75.0 & 75.3 & 75.3 & 75.3 & 75.3 \\ 
        rain            & 75.4 & 72.0 & 61.7 & 23.9 & 7.1  & 2.1 \\ 
        snow            & 75.4 & 59.6 & 57.4 & 55.9 & 58.1 & 60.8 \\ 
        fog             & 75.4 & 39.3 & 21.0 & 19.7 & 8.1  & 2.4 \\
        \hline
        Static          & 75.4 & 63.1 & 56.2 & 47.4 & 31.4 & 12.7 \\
        \specialrule{1.5pt}{0pt}{0pt}
    \end{tabular}%
    }
    \caption{SwinGait robustness scores on CASIA-B dataset (rounded to one decimal place).}
    \label{tab:swingaitcasia}
\end{table}

\begin{table}[t!]
    \centering
    \setlength{\tabcolsep}{4pt} 
    \resizebox{\linewidth}{!}{%
    \begin{tabular}{l|cccccc}
        \specialrule{1.5pt}{0pt}{0pt}
        \rowcolor{mygray}
        \textbf{Perturbations} & \textbf{Clean} & \textbf{Sev 1} & \textbf{Sev 2} & \textbf{Sev 3} & \textbf{Sev 4} & \textbf{Sev 5} \\ 
        \hline\hline
        gaussian\_noise & 82.1 & 82.1 & 79.0 & 60.3 & 37.8 & 31.9 \\
        defocus\_blur   & 82.1 & 31.6 & 31.3 & 31.2 & 31.2 & 31.1 \\
        impulse\_noise  & 82.1 & 82.1 & 79.4 & 50.7 & 36.1 & 31.2 \\
        speckle\_noise  & 82.1 & 38.1 & 35.1 & 32.9 & 31.7 & 31.3 \\
        shot\_noise     & 82.1 & 45.5 & 34.4 & 32.0 & 31.3 & 31.2 \\
        motion\_blur    & 82.1 & 66.5 & 63.8 & 59.2 & 58.5 & 59.5 \\
        zoom\_in        & 82.1 & 72.5 & 62.5 & 50.8 & 49.9 & 61.0 \\
        \hline
        freeze          & 82.1 & 58.5 & 53.2 & 48.1 & 43.1 & 48.0 \\
        \hline
        snow            & 82.1 & 56.6 & 55.7 & 54.7 & 54.3 & 54.2 \\
        fog             & 82.1 & 65.8 & 64.0 & 62.0 & 60.7 & 56.6 \\
        \hline
        Static          & 82.1 & 52.6 & 50.4 & 42.0 & 36.6 & 35.4 \\
        \specialrule{1.5pt}{0pt}{0pt}
    \end{tabular}%
    }
    \caption{DeepGait robustness scores on CCPG dataset (rounded to one decimal place).}
    \label{tab:deepgaitccpg}
\end{table}

\begin{table}[t!]
    \centering
    \setlength{\tabcolsep}{4pt} 
    \resizebox{\linewidth}{!}{%
    \begin{tabular}{l|cccccc}
        \specialrule{1.5pt}{0pt}{0pt}
        \rowcolor{mygray}
        \textbf{Perturbations} & \textbf{Clean} & \textbf{Sev 1} & \textbf{Sev 2} & \textbf{Sev 3} & \textbf{Sev 4} & \textbf{Sev 5} \\ 
        \hline\hline
        gaussian\_noise & 77.4 & 77.4 & 74.5 & 56.7 & 35.5 & 29.9 \\
        defocus\_blur   & 77.4 & 29.8 & 29.6 & 29.5 & 29.4 & 29.1 \\
        impulse\_noise  & 77.4 & 77.3 & 74.9 & 47.6 & 34.0 & 29.3 \\
        speckle\_noise  & 77.4 & 35.9 & 33.1 & 30.9 & 29.8 & 29.6 \\
        shot\_noise     & 77.4 & 42.7 & 32.4 & 30.0 & 29.0 & 29.3 \\
        motion\_blur    & 77.4 & 57.4 & 54.5 & 50.5 & 51.8 & 54.6 \\
        zoom\_in        & 77.4 & 67.8 & 57.4 & 43.3 & 39.5 & 49.0 \\
        \hline
        freeze          & 77.4 & 54.3 & 50.4 & 45.0 & 39.7 & 44.7 \\
        \hline
        snow            & 77.4 & 48.8 & 47.9 & 47.1 & 46.6 & 46.3 \\
        fog             & 77.4 & 57.0 & 55.2 & 53.6 & 52.6 & 48.4 \\
        \hline
        Static          & 77.4 & 45.4 & 43.7 & 36.6 & 33.4 & 33.6 \\
        \specialrule{1.5pt}{0pt}{0pt}
    \end{tabular}%
    }
    \caption{GaitBase robustness scores on CCPG dataset (rounded to one decimal place).}
    \label{tab:gaitbaseccpg}
\end{table}

\begin{table}[t!]
    \centering
    \setlength{\tabcolsep}{4pt} 
    \resizebox{\linewidth}{!}{%
    \begin{tabular}{l|cccccc}
        \specialrule{1.5pt}{0pt}{0pt}
        \rowcolor{mygray}
        \textbf{Perturbations} & \textbf{Clean} & \textbf{Sev 1} & \textbf{Sev 2} & \textbf{Sev 3} & \textbf{Sev 4} & \textbf{Sev 5} \\ 
        \hline\hline
        gaussian\_noise & 70.2 & 26.9 & 31.8 & 51.2 & 67.6 & 70.2 \\
        defocus\_blur   & 70.2 & 26.2 & 26.3 & 26.4 & 26.5 & 26.7 \\
        impulse\_noise  & 70.2 & 26.3 & 30.6 & 42.9 & 67.9 & 70.2 \\
        speckle\_noise  & 70.2 & 26.4 & 26.7 & 27.7 & 29.5 & 32.1 \\
        shot\_noise     & 70.2 & 26.3 & 26.3 & 26.5 & 29.0 & 38.5 \\
        motion\_blur    & 70.2 & 56.6 & 54.4 & 50.9 & 51.0 & 51.4 \\
        zoom\_in        & 70.2 & 53.4 & 45.5 & 43.6 & 54.7 & 62.5 \\
        \hline
        freeze          & 70.2 & 53.7 & 51.3 & 47.4 & 42.7 & 47.1 \\
        \hline
        snow            & 70.2 & 49.2 & 48.4 & 47.6 & 47.3 & 47.1 \\
        fog             & 70.2 & 55.8 & 54.3 & 52.8 & 51.7 & 47.7 \\
        \hline
        Static          & 70.2 & 44.4 & 43.8 & 35.1 & 30.8 & 29.9 \\
        \specialrule{1.5pt}{0pt}{0pt}
    \end{tabular}%
    }
    \caption{SwinGait robustness scores on CCPG dataset (rounded to one decimal place).}
    \label{tab:swingaitccpg}
\end{table}

\begin{table*}[t!]
    \centering
    \begin{minipage}{0.48\linewidth}
        \centering
        \begin{tabular}{l|cccc}
            \specialrule{1.5pt}{0pt}{0pt}
            \rowcolor{mygray}
            \textbf{Perturbations} & \textbf{Clean} & \textbf{Sev 1} & \textbf{Sev 3} & \textbf{Sev 5} \\
            \hline\hline
            gaussian\_noise & 75.9 & 7.3 & 6.4 & 3.2 \\
            impulse\_noise & 75.9 & 6.4 & 5.3 & 3.8 \\
            speckle\_noise & 75.9 & 12.3 & 14.1 & 15.7 \\
            motion\_blur & 75.9 & 64.2 & 27.4 & 5.7 \\
            \hline
            freeze & 75.9 & 68.6 & 51.4 & 52.2 \\
            \hline
            rain & 75.9 & 69.0 & 41.5 & 9.4 \\
            snow & 75.9 & 66.2 & 64.5 & 62.4 \\
            fog & 75.9 & 77.8 & 76.0 & 69.2 \\
            \hline
            Static & 75.9 & 50.1 & 26.0 & 1.2 \\
            \specialrule{1.5pt}{0pt}{0pt}
        \end{tabular}
    \end{minipage}
    \hfill
    \begin{minipage}{0.48\linewidth}
        \centering
        \begin{tabular}{l|cccc}
            \specialrule{1.5pt}{0pt}{0pt}
            \rowcolor{mygray}
            \textbf{Perturbations} & \textbf{Clean} & \textbf{Sev 1} & \textbf{Sev 3} & \textbf{Sev 5} \\
            \hline\hline
            gaussian\_noise & 65.5 & 5.2 & 5.3 & 2.8 \\
            impulse\_noise & 65.5 & 3.7 & 3.8 & 3.4 \\
            speckle\_noise & 65.5 & 9.1 & 12.4 & 13.7 \\
            motion\_blur & 65.5 & 59.4 & 24.6 & 4.7 \\
            \hline
            freeze & 65.5 & 62.0 & 50.1 & 51.0 \\
            \hline
            rain & 65.5 & 65.5 & 39.3 & 8.1 \\
            snow & 65.5 & 60.9 & 59.7 & 58.1 \\
            fog & 65.5 & 68.2 & 66.4 & 61.5 \\
            \hline
            Static & 65.5 & 35.9 & 17.7 & 1.1 \\
            \specialrule{1.5pt}{0pt}{0pt}
        \end{tabular}
    \end{minipage}
    \caption{Robustness scores of \textbf{DeepGait} (left) and \textbf{SwinGait} (right) on the \textbf{SUSTech1K} dataset across selected perturbation severities.}
    \label{tab:dg_sg_sustech_side}
\end{table*}

\begin{table}[t!]
    \centering
    \setlength{\tabcolsep}{5pt} 
    \begin{tabular}{l|cccc}
        \specialrule{1.5pt}{0pt}{0pt}
        \rowcolor{mygray}
        \textbf{Perturbations} & \textbf{Clean} & \textbf{Sev 1}  & \textbf{Sev 3} &  \textbf{Sev 5} \\ 
        \hline\hline
        gaussian\_noise & 83.6 & 9.6 & 9.4 & 4.0 \\
        impulse\_noise & 83.6 & 6.7 & 5.3 & 3.8 \\
        speckle\_noise & 83.6 & 14.0 & 15.2 & 16.3 \\
        motion\_blur & 83.6 & 71.9 & 33.6 & 5.6 \\
        \hline
        freeze & 83.6 & 81.5 & 70.8 & 71.4 \\
        \hline
        rain & 83.6 & 76.0 & 48.2 & 9.8 \\
        snow & 83.6 & 73.7 & 72.5 & 71.2 \\
        \hline
        Static & 83.6 & 51.7 & 26.9 & 1.4 \\
        \specialrule{1.5pt}{0pt}{0pt}
    \end{tabular}
    \caption{GaitBase Robustness Scores on SUSTech1K dataset.}
    \label{tab:gaitbasesustech}
\end{table}

\textbf{Choice of Segmentation Model}
We use the SCHP model for our analysis as it offers the best trade-off in computational efficiency compared to other segmentation and parsing backbones.
\begin{table*}[t!]
    \centering
    \setlength{\tabcolsep}{4pt} % adjust column spacing
    \renewcommand{\arraystretch}{1.1} % row height
    \begin{tabular}{lcccc}
        \toprule
        \rowcolor{mygray}
        \textbf{Model} & \textbf{Params} & \textbf{GFLOPs} & \textbf{Inference Time} & \textbf{FPS} \\
        \midrule
        SCHP\textcolor{lightgray}{\scriptsize{[TPAMI20]}}  & 66.7M  & 87.36   & 43.75  & 22.86 \\
        CDGNet\textcolor{lightgray}{\scriptsize{[CVPR22]}}  & 80.9M  & 162.86  & 47.71  & 20.96 \\
        M2FP\textcolor{lightgray}{\scriptsize{[arXiv23]}}   & 63.0M  & 92.73   & 78.45  & 12.75 \\
        GSAM\textcolor{lightgray}{\scriptsize{[arXiv24]}}   & 874M   & 2984.06 & 865.99 & 1.15 \\
        \bottomrule 
    \end{tabular}
    \caption{Comparison of model complexity and efficiency.}
    \label{tab:model_efficiency}
\end{table*}

\textbf{Analysis on Noise Severity}
We show the detailed analysis on how severity of noise impacts silhouettes and model features.
\begin{figure*}[t!]
    \centering
    \includegraphics[width=0.9\linewidth]{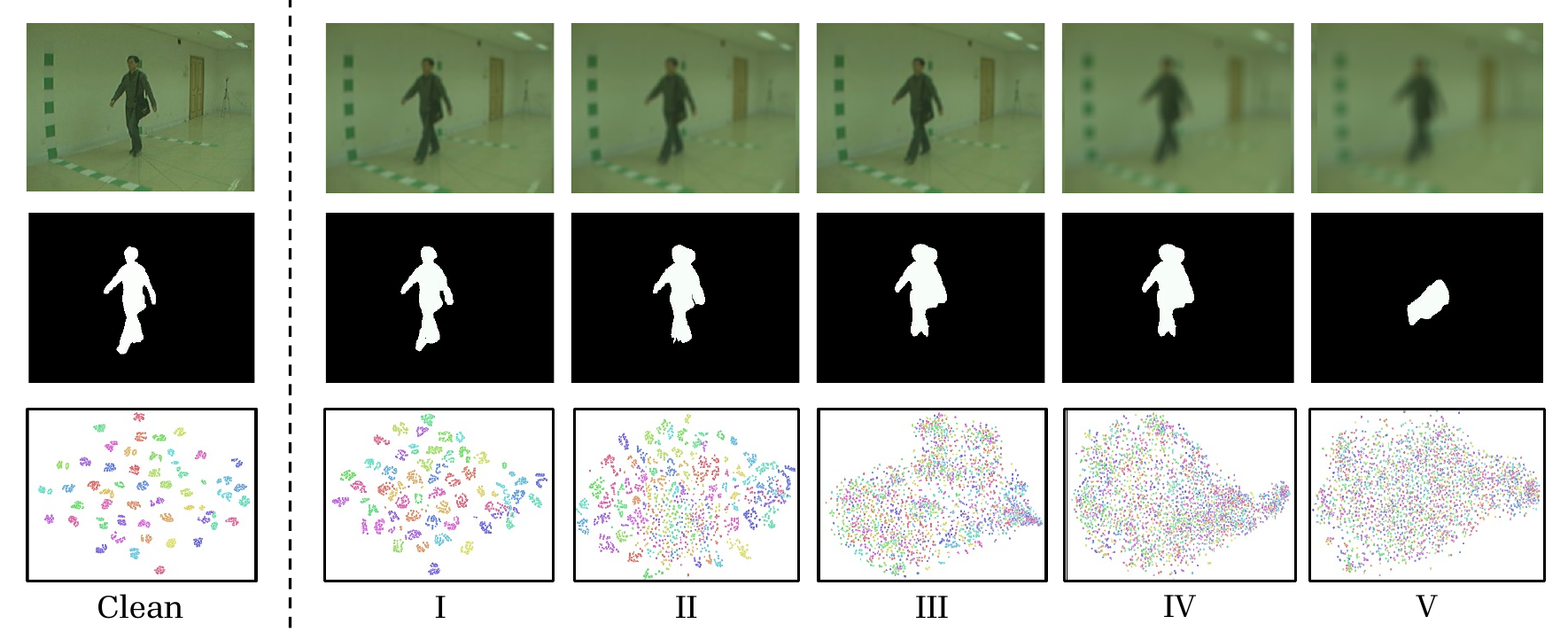}
    \caption{\textbf{Qualitative Analysis} of increasing digital noise severity on CASIA-B. 
    \textit{(Top)} Silhouettes from the parsing model degrade visibly with higher noise, losing structural integrity. 
    \textit{(Bottom)} t-SNE of DeepGait features shows reduced cluster separability, leading to weakened identity discrimination.}
    \label{fig:tsne}
\end{figure*}

\textbf{Robustness Analysis with Noisy Gallery:} We construct a \textit{fixed noisy gallery} by applying one of the 15 corruption types to each gallery sequence, with severity levels sampled randomly using the probabilities 0.6 (severity 1), 0.3 (severity 2), and 0.1 (severity 3). This noisy gallery is held constant across evaluations. We then evaluate each gait recognition model using 15 perturbed probe sets—each corresponding to one of the 15 corruption types—against this shared noisy gallery.
Table~\ref{tab:noisy_gallery_casiab} reports the absolute ($\delta_a$) and relative ($\delta_r$) robustness scores for each model on the CASIA-B dataset across four corruption categories. 

\begin{table*}[t!]
    \centering
    \renewcommand{\arraystretch}{1.06}
    \begin{tabular}{r|cc|cc|cc|cc}
        \specialrule{1.5pt}{0pt}{0pt}
        \rowcolor{mygray} 
        \textbf{Method} & \multicolumn{2}{c|}{\textbf{Camera}} & \multicolumn{2}{c|}{\textbf{Temporal}} & \multicolumn{2}{c|}{\textbf{Environmental}} & \multicolumn{2}{c}{\textbf{Occlusion}} \\
        \rowcolor{mygray}
        & $\delta_a$ & $\delta_r$ & $\delta_a$ & $\delta_r$ & $\delta_a$ & $\delta_r$ & $\delta_a$ & $\delta_r$ \\
        \hline\hline
        DeepGaitV2\textcolor{lightgray}{\scriptsize{[arXiv23]}} & 0.36 & 0.26 & 0.75 & 0.71 & 0.45 & 0.37 & 0.56 & 0.49 \\
        GaitGL\textcolor{lightgray}{\scriptsize{[ICCV21]}} & 0.32 & 0.22 & 0.72 & 0.67 & 0.39 & 0.29 & 0.49 & 0.40 \\
        GaitBase\textcolor{lightgray}{\scriptsize{[CVPR23]}} & 0.33 & 0.21 & 0.77 & 0.73 & 0.40 & 0.30 & 0.46 & 0.37 \\
        GaitSet\textcolor{lightgray}{\scriptsize{[AAAI19]}} & 0.42 & 0.24 & 0.83 & 0.79 & 0.45 & 0.29 & 0.55 & 0.41 \\
        GaitPart\textcolor{lightgray}{\scriptsize{[CVPR20]}} & 0.39 & 0.25 & 0.84 & 0.81 & 0.42 & 0.29 & 0.52 & 0.41 \\
        SwinGait\textcolor{lightgray}{\scriptsize{[arXiv23]}} & \textbf{0.46} & \textbf{0.29} & \textbf{0.86} & \textbf{0.82} & \textbf{0.53} & \textbf{0.37} & \textbf{0.62} & \textbf{0.50} \\
        \specialrule{1.5pt}{0pt}{0pt}
    \end{tabular}
    \caption{Absolute ($\delta_a$) and relative ($\delta_r$) robustness scores of gait models on CASIA-B with a fixed noisy gallery.}
    \label{tab:noisy_gallery_casiab}
\end{table*}

\textbf{Robustness via Noise-Aware Training Details:} We create a perturbed training set with five representative corruption types, covering camera viewpoint changes, temporal distortions, environmental artifacts, and occlusions. These perturbations are applied with severity levels 1, 2, and 3, sampled according to a probability distribution of 0.6, 0.3, and 0.1, respectively. The remaining ten corruption types, which are unseen during training, are used to generate the noisy test set for evaluation. We train gait recognition models with varying ratios of clean and noisy training data (i.e., 100:0, 80:20, 50:50, and 20:80), and evaluate them on both the original clean test set and the noisy test set. The performance under these settings is summarized in Table~\ref{tab:aug_noisy_test} and Table~\ref{tab:aug_clean_test}.

\begin{table*}[t!]
    \centering
    \renewcommand{\arraystretch}{1.2}
    \setlength{\tabcolsep}{5.2pt}
    \begin{tabular}{r|ccc|ccc|ccc}
        \specialrule{1.5pt}{0pt}{0pt}
        \rowcolor{mygray}
        \textbf{Noise Ratio} & \multicolumn{3}{c|}{\textbf{GaitBase\textcolor{lightgray}{\scriptsize{[CVPR23]}}}} & \multicolumn{3}{c|}{\textbf{DeepGaitV2\textcolor{lightgray}{\scriptsize{[Arxiv23]}}}} & \multicolumn{3}{c}{\textbf{SwinGait\textcolor{lightgray}{\scriptsize{[Arxiv23]}}}} \\
        \rowcolor{mygray}
        & NM & BG & CL & NM & BG & CL & NM & BG & CL \\
        \hline \hline
        No Noise & 69.34 & 60.30 & 44.92 & 69.70 & 60.04 & 45.55 & 67.18 & 56.06 & 39.42 \\
        20\% & 73.06 & 64.13 & 46.31 & 75.07 & 64.06 & 46.80 & 71.16 & 59.76 & 39.29 \\
        50\% & 73.84 & 64.34 & 44.87 & 75.29 & 64.27 & 44.43 & 71.48 & 57.81 & 37.72 \\
        80\% & 73.69 & 63.40 & 43.03 & 74.43 & 61.64 & 42.20 & 71.40 & 56.47 & 34.53 \\
        \specialrule{1.5pt}{0pt}{0pt}
    \end{tabular}
    \caption{Rank-1 accuracy (\%) on the noisy test set (CASIA-B) across different training noise ratios.}
    \label{tab:aug_noisy_test}
\end{table*}

\begin{table*}[t!]
    \centering
    \label{tab:clean_test_pivot}
    \renewcommand{\arraystretch}{1.2}
    \setlength{\tabcolsep}{5.2pt}
    \begin{tabular}{r|ccc|ccc|ccc}
        \specialrule{1.5pt}{0pt}{0pt}
        \rowcolor{mygray}
        \cellcolor{mygray} \textbf{Noise Ratio} & \multicolumn{3}{c|}{\textbf{GaitBase}\textcolor{lightgray}{\scriptsize{[CVPR23]}}} & \multicolumn{3}{c|}{\textbf{DeepGaitV2}\textcolor{lightgray}{\scriptsize{[Arxiv23]}}} & \multicolumn{3}{c}{\textbf{SwinGait}\textcolor{lightgray}{\scriptsize{[Arxiv23]}}} \\
        \rowcolor{mygray}
        \cellcolor{mygray} & NM & BG & CL & NM & BG & CL & NM & BG & CL \\
        \hline\hline
        No Noise & 95.66 & 90.40 & 74.88 & 94.93 & 89.60 & 74.97 & 90.64 & 82.52 & 63.17 \\
        20\% & 94.62 & 88.23 & 72.20 & 94.43 & 86.78 & 69.67 & 90.14 & 80.48 & 57.55 \\
        50\% & 93.23 & 85.62 & 66.98 & 91.35 & 84.75 & 64.01 & 89.40 & 77.82 & 53.56 \\
        80\% & 91.83 & 83.42 & 63.24 & 90.98 & 81.22 & 61.49 & 88.65 & 74.63 & 49.62 \\
        \specialrule{1.5pt}{0pt}{0pt}
    \end{tabular}
    \caption{Rank-1 accuracy (\%) on the original clean test set (CASIA-B) across different training noise ratios.}
    \label{tab:aug_clean_test}
\end{table*}

\section{Training Details: Efficient Distillation}
\label{sec:distillation}

We implement a two-stream distillation framework to improve robustness against noise in gait embeddings. The training procedure involves a fixed teacher network and a learnable student network. The teacher is applied only to clean sequences, while the student is trained on both clean and noisy inputs.
We use a contrastive loss between the teacher's embedding (extracted from clean inputs) and the student's embeddings (from both clean and noisy sequences) to align the representational space. Let $E_T(x)$ and $E_S(x)$ denote the embeddings from the teacher and student, respectively. The contrastive losses are defined as:
\begin{align}
\mathcal{L}_{\text{con}}^{\text{clean}} &= 
   \text{Con}(E_T(x_{\text{clean}}), E_S(x_{\text{clean}})), \notag \\
\mathcal{L}_{\text{con}}^{\text{noisy}} &= 
   \text{Con}(E_T(x_{\text{clean}}), E_S(x_{\text{noisy}})).
\end{align}

where $\text{Con}(\cdot, \cdot)$ denotes a normalized temperature-scaled contrastive loss.

Additionally, the student is trained with softmax and triplet losses using both clean and noisy inputs. The softmax losses are given by
\begin{align}
\mathcal{L}_{\text{softmax}}^{\text{clean}} &=
   \text{CE}(\mathbf{z}_{S}^{\text{clean}}, y), \\
\mathcal{L}_{\text{softmax}}^{\text{noisy}} &=
   \text{CE}(\mathbf{z}_{S}^{\text{noisy}}, y).
\end{align}
and the triplet losses are defined as
\begin{align}
\mathcal{L}_{\text{triplet}}^{\text{clean}} &=
   \text{Triplet}(E_S(x_{\text{clean}}), y), \\
\mathcal{L}_{\text{triplet}}^{\text{noisy}} &=
   \text{Triplet}(E_S(x_{\text{noisy}}), y).
\end{align}

\begin{align}
\mathcal{L}_{\text{total}} &=
   \lambda_1 \mathcal{L}_{\text{con}}^{\text{clean}} 
 + \lambda_2 \mathcal{L}_{\text{con}}^{\text{noisy}} 
 + \lambda_3 \mathcal{L}_{\text{softmax}}^{\text{clean}} \notag \\
&\quad
 + \lambda_4 \mathcal{L}_{\text{softmax}}^{\text{noisy}}
 + \lambda_5 \mathcal{L}_{\text{triplet}}^{\text{clean}}
 + \lambda_6 \mathcal{L}_{\text{triplet}}^{\text{noisy}} .
\end{align}

During inference, only the student model is used.

\section{Scaling to Large Real-World Dataset MEVID}
\label{sec:mevid}
To assess model robustness and scalability in unconstrained settings, we evaluate on the MEVID dataset. MEVID is a large-scale video-based re-identification benchmark comprising 8,092 tracklets of 158 subjects recorded across 73 days in 33 camera views spanning 17 locations. Each subject appears in multiple sessions with varied clothing (598 outfits in total), motion styles, and environments (indoor/outdoor), making it well-suited for testing gait models under real-world challenges. The tracklets average 590 frames each and include significant variations in background clutter, occlusion, lighting, and viewpoints. MEVID ensures diversity in geography and activity context. We train each model from scratch on the MEVID training set using standard classification and metric learning objectives.
\begin{table*}[t!]
    \centering
    \renewcommand{\arraystretch}{1.1}
    \scalebox{0.93}{
    \setlength{\tabcolsep}{6pt}
    \begin{tabular}{r|c|c|c|c|c}
        \specialrule{1.5pt}{0pt}{0pt}
        \rowcolor{mygray}
        \textbf{Model} & \textbf{mAP} & \textbf{Top-1} & \textbf{Top-5} & \textbf{Top-10} & \textbf{Top-20} \\
        \hline\hline
        GaitBase\textcolor{lightgray}{\scriptsize{[CVPR23]}} 
            & 11.5 & \textbf{22.5} & \textbf{34.3} & 41.6 & 47.0 \\
        GaitGL\textcolor{lightgray}{\scriptsize{[ICCV21]}}
            & 7.1  & 6.0  & 16.5 & 23.8 & 32.4 \\
        DeepGaitV2\textcolor{lightgray}{\scriptsize{[Arxiv23]}}
            & 8.7  & 16.5 & 30.2 & 36.2 & 45.1 \\
        SwinGait\textcolor{lightgray}{\scriptsize{[Arxiv23]}}
            & \textbf{11.7} & 16.2 & 29.5 & 39.7 & \textbf{49.2} \\
        GaitSet\textcolor{lightgray}{\scriptsize{[AAAI19]}}
            & 9.6 & 8.9  & 27.9 & 38.1 & 48.9 \\
        GaitPart\textcolor{lightgray}{\scriptsize{[CVPR20]}}
            & 9.1 & 13.3 & 27.6 & 36.8 & 49.1 \\
        \specialrule{1.5pt}{0pt}{0pt}
    \end{tabular}
    }
    \caption{Training performance of gait models on the MEVID dataset. We report mAP and Top-1/5/10/20 retrieval accuracy (\%).}
    \label{tab:mevid_training}
\end{table*}

\section{Details on Corruptions}
\label{sec:details}
% Details for training implementation of noises

We present the details of each and every noises and how it is implemented.

\textbf{Gaussian Noise}
The Gaussian Noise function introduces Gaussian-distributed noise to each frame in an array of video frames. The noise severity is determined by a predefined scale corresponding to different severity levels: 0.08, 0.12, 0.18, 0.26, and 0.38. Each frame is normalized to a [0, 1] range before noise addition. After applying the noise, the pixel values are clipped to ensure they remain within the [0, 1] range, and then the frames are rescaled back to [0, 255]. Finally, the processed frames are converted back to an unsigned 8-bit integer format.

\textbf{Speckle Noise}
Each frame is normalized to a [0, 1] range before noise addition. Speckle noise is generated by multiplying the normalized image by Gaussian noise scaled by predefined severity levels: 0.15, 0.2, 0.25, 0.3, and 0.35. The np.random.normal function generates Gaussian-distributed noise, which is added to each pixel of the frame. The noisy image is then clipped to the [0, 1] range and rescaled back to [0, 255], before being converted to an unsigned 8-bit integer format.

\textbf{Shot Noise}
Each frame is normalized to a [0, 1] range before noise addition. The noisy image is generated by scaling the normalized image by predefined severity levels: 250, 100, 50, 30, and 15, then passed to the np.random.poisson function, which adds Poisson-distributed noise. This noise models the random variation of photon count in low-light conditions. The resulting image is then divided by the severity level value to normalize it. The noisy image is clipped to the [0, 1] range and rescaled back to [0, 255], before being converted to an unsigned 8-bit integer format.

\textbf{Impulse Noise}
Each frame is normalized to a [0, 1] range before noise addition. Salt-and-pepper noise is introduced using the skimage.util.random\_noise function in 's\&p' mode. This function randomly sets a proportion of pixels to either 0 or 1, based on severity levels: 0.03, 0.06, 0.09, 0.17, and 0.27. The noisy image is then clipped to the [0, 1] range and rescaled back to [0, 255], before being converted to an unsigned 8-bit integer format.

\textbf{Defocus Blur}
Each frame is normalized to a [0, 1] range before blur addition. A disk-shaped kernel is created using the disk function, with radius values based on severity levels: 3, 4, 6, 8, and 10, and an alias blur of 0.1 to 0.5. The kernel simulates out-of-focus blur and is applied to each color channel of the frame using cv2.filter2D. The blurred image is clipped to the [0, 1] range and rescaled back to [0, 255], before being converted to an unsigned 8-bit integer format.

\textbf{Zoom Blur}
Each frame is normalized to a [0, 1] range before blur addition. The frames are zoomed by factors defined by severity levels: 1-1.11, 1-1.16, 1-1.21, 1-1.26, and 1-1.31. This is done using the scipy.ndimage.zoom function, which interpolates the image to create a zoom effect. For each severity level, a range of zoom factors is applied, and the resulting images are averaged to create a smooth blur effect. Specifically, scipy.ndimage.zoom is used to resample the image at different scales. Finally, the frames are rescaled back to [0, 255] and converted to an unsigned 8-bit integer format.

\textbf{Motion Blur}
The motion blur function simulates motion in video frames by applying a motion blur effect using the Wand library's MagickMotionBlurImage function. Each frame is converted to the WandImage format and processed with a motion blur effect defined by varying radii and sigma values: (10, 3), (15, 5), (15, 8), (15, 12), and (20, 15). The motion blur method uses these parameters to create a directional blur, simulating motion at different speeds and angles. The frames are then converted back to numpy arrays, clipped to the [0, 255] range, and returned as unsigned 8-bit integers.

\textbf{Zoom In}
The zoom in function simulates a gradual zoom-in effect on each frame. Zoom factors are determined by predefined severity levels: 1.5, 2.0, 2.5, 3.0, and 3.5. For each frame, a zoom matrix is created using cv2.getRotationMatrix2D, which specifies the center of the zoom and the scaling factor. The cv2.warpAffine function is then used to apply this transformation to the frame, effectively zooming in. The frames are processed incrementally, creating a smooth zoom effect over time. Finally, the processed frames are normalized to [0, 255] and converted to an unsigned 8-bit integer format.

\textbf{Freeze}
The freeze function mimics the effect of a frame freeze by randomly selecting and repeating certain frames. The severity levels determine the proportion of frames to be repeated: 40\%, 20\%, 10\%, 5\%, and 10\%. The function ensures the transition between frozen and regular frames is smooth by duplicating the selected frames in a way that the sequence of repeated frames appears natural. This is achieved by selecting the frames at random intervals and ensuring that the duplicates are seamlessly integrated with the regular frames. After processing, the frames are clipped to [0, 255] and converted to an unsigned 8-bit integer format.

\textbf{Sampling}
The sampling function reduces the frame rate by downsampling and then upsampling the frames. Severity levels define the downsampling rates: 2, 4, 8, 16, and 32. The frames are first downsampled by selecting every nth frame and then upsampled by repeating these frames to match the original frame count. This mimics the effect of a lower frame rate, simulating scenarios with limited bandwidth or processing power. The processed frames are clipped to [0, 255] and converted to an unsigned 8-bit integer format.

\textbf{Low Light}
The low light function applies a vignette effect to simulate low light conditions. The severity levels, defined by vignette strength: 1, 2, 3, 4, and 5, determine the darkness of the edges. A mask is created using np.mgrid to simulate a light source effect, darkening the edges while keeping the center bright. This mask is applied to each frame, adjusting the brightness to create a realistic low-light environment. The vignette mask decreases linearly from the center to the edges, simulating the effect of a light source fading out. Finally, the frames are clipped to [0, 255] and converted to an unsigned 8-bit integer format.

\textbf{Fog}
The fog function uses the Albumentations library to simulate fog effects in video frames. The severity of the fog is controlled by fog coefficients, which determine the density and intensity of the fog. These coefficients are predefined for each severity level: 0.49, 0.59, 0.69, 0.79, and 0.89. The A.RandomFog function is employed, which creates a fog effect by overlaying a semi-transparent white layer over the image, reducing the contrast and adding a hazy appearance. The alpha\_coef parameter controls the transparency of the fog, while fog\_coef\_lower and fog\_coef\_upper set the range for the fog density.

\textbf{Rain}
The rain function also utilizes the Albumentations library to overlay realistic rain streaks on video frames. The severity of the rain is defined by rain types and parameters: "drizzle", "drizzle", None, "heavy", and "torrential", with corresponding brightness coefficients (0.7, 0.7, 0.6, 0.55, and 0.5) and drop lengths (5, 15, 20, 40, and 50). The A.RandomRain function simulates rain by adding streaks and adjusting brightness. The parameters slant\_lower and slant\_upper set the angle of the rain streaks, drop\_length controls the length of each streak, and blur\_value determines the blurriness of the rain. The brightness\_coefficient adjusts the brightness of the image to simulate the darkening effect of rain. The rain effect is applied to each frame, creating a consistent simulation of rainfall. Finally, the processed frames are converted back to an unsigned 8-bit integer format.

\textbf{Snow}
The snow function adds snowflakes and increases brightness using the Albumentations library. The severity of the snow is controlled by snow coefficients: 0.05, 0.1, 0.15, 0.2, and 0.25. The A.RandomSnow function is employed to simulate snow by overlaying white noise on the image and increasing the brightness to mimic the reflective nature of snow. Parameters like snow\_point and brightnes\_coeff are adjusted to control the density and intensity of the snowfall. The brightness\_coefficient increases the overall brightness to simulate the glare and reflection caused by snow. Finally, the processed frames are converted back to an unsigned 8-bit integer format.

\textbf{Occlusion}
The occlusion function introduces random obstructions in video frames by overlaying object masks from the COCO dataset. The severity of the occlusion is determined by the extent to which the object covers the frame. Image IDs in the COCO dataset are sorted by the area occupied by objects, and this sorted list is divided into five groups based on severity. Depending on the severity level, a group is selected, and a random object from this group is used. The corresponding mask is retrieved using coco.annToMask, resized to fit the frame dimensions, and applied using PIL.Image.paste, blending the masked objects into the scene. This process simulates partial occlusions by static objects

% {
%     \small
%     \bibliographystyle{ieeenat_fullname}
%     \bibliography{main}
% }